\pdfoutput=1
\PassOptionsToPackage{table}{xcolor}
\documentclass[11pt]{article}

\usepackage[preprint]{acl}

\usepackage[T1]{fontenc}     
\usepackage[utf8]{inputenc} 
\usepackage{times}
\usepackage{latexsym}
\usepackage{inconsolata}    
\usepackage{microtype}      

\usepackage{amsmath}
\usepackage{amsfonts}
\usepackage{bbm}
\usepackage{graphicx}
\usepackage{lipsum}
\usepackage{multicol}
\usepackage{multirow}
\usepackage{booktabs}
\usepackage{footmisc}
\usepackage{footnote}
\usepackage{enumitem}
\usepackage{pifont}
\usepackage{subcaption}
\usepackage{worldflags}
\usepackage{url}
\usepackage{xurl}
\usepackage{float}
\usepackage{rotating}
\usepackage{tablefootnote}
\usepackage{arydshln}
\usepackage{tabularx}
\usepackage{tcolorbox}

\definecolor{tab-lightgrey}{HTML}{F4F4F4}
\definecolor{tab-darkgrey}{HTML}{EEEEEE}

\hypersetup{
  pdftitle={UniversalCEFR: Enabling Open Multilingual Research on Language Proficiency Assessment},
  pdfauthor={Joseph Marvin Imperial, Abdullah Barayan, Regina Stodden, Rodrigo Wilkens, Ricardo Munoz Sanchez, Lingyun Gao, Melissa Torgbi, Dawn Knight, Gail Forey, Reka R. Jablonkai, Ekaterina Kochmar, Robert Reynolds, Eugenio Ribeiro, Horacio Saggion, Elena Volodina, Sowmya Vajjala, Thomas Francois, Fernando Alva-Manchego, Harish Tayyar Madabushi},
  colorlinks=true,
  linkcolor=blue,
  urlcolor=blue
}

\title{\textsc{UniversalCEFR}: Enabling Open Multilingual Research on Language Proficiency Assessment}

\author{
  \textbf{Joseph Marvin Imperial}\textnormal{\textsuperscript{1,3}},
  \textbf{Abdullah Barayan}\textnormal{\textsuperscript{2,14}},
  \textbf{Regina Stodden}\textnormal{\textsuperscript{4}},
\\
  \textbf{Rodrigo Wilkens}\textsuperscript{5},
  \textbf{Ricardo Muñoz Sánchez}\textsuperscript{6},
  \textbf{Lingyun Gao}\textsuperscript{7},
  \textbf{Melissa Torgbi}\textsuperscript{1},
\\
  \textbf{Dawn Knight}\textsuperscript{2},
  \textbf{Gail Forey}\textsuperscript{1},
  \textbf{Reka R. Jablonkai}\textsuperscript{1},
  \textbf{Ekaterina Kochmar}\textsuperscript{8},
\\
 \textbf{Robert Reynolds}\textsuperscript{9},
  \textbf{Eugénio Ribeiro}\textsuperscript{10,11},
  \textbf{Horacio Saggion}\textsuperscript{12},
\\
  \textbf{Elena Volodina}\textsuperscript{6},
  \textbf{Sowmya Vajjala}\textsuperscript{13},
 \textbf{Thomas François}\textsuperscript{7},
\\
  \textbf{Fernando Alva-Manchego}\textsuperscript{2},
  \textbf{Harish Tayyar Madabushi}\textsuperscript{1}
\\
  \textsuperscript{1}University of Bath,
  \textsuperscript{2}Cardiff University,
  \textsuperscript{3}National University Philippines,\\
  \textsuperscript{4}Bielefeld University,
  \textsuperscript{5}University of Exeter,
  \textsuperscript{6}University of Gothenburg,
  \textsuperscript{7}UCLouvain,\\
  \textsuperscript{8}MBZUAI,
  \textsuperscript{9}Brigham Young University,
  \textsuperscript{10}INESC-ID Lisboa,\\
  \textsuperscript{11}Instituto Universitário de Lisboa (ISCTE-IUL),
  \textsuperscript{12}Universitat Pompeu Fabra,\\
  \textsuperscript{13}National Research Council, Canada,
 \textsuperscript{14}King Abdulaziz University\\
  \small{
  \textsc{Correspondence:}
    \texttt{\href{mailto:jmri20@bath.ac.uk}{jmri20@bath.ac.uk}}, \texttt{\href{mailto:alvamanchegof@cardiff.ac.uk}{alvamanchegof@cardiff.ac.uk}}
  }
  \hspace*{0.15cm}
}

\begin{document}
\maketitle
\begin{abstract}
We introduce \textbf{\textsc{UniversalCEFR}}, a large-scale multilingual and multidimensional dataset of texts annotated with CEFR (Common European Framework of Reference) levels in 13 languages. To enable open research in automated readability and language proficiency assessment, \textsc{UniversalCEFR} comprises \textbf{505,807 CEFR-labeled texts} curated from educational and learner-oriented resources, standardized into a unified data format to support consistent processing, analysis, and modelling across tasks and languages. To demonstrate its utility, we conduct benchmarking experiments using three modelling paradigms: a) linguistic feature-based classification, b) fine-tuning pre-trained LLMs, and c) descriptor-based prompting of instruction-tuned LLMs. Our results support using linguistic features and fine-tuning pretrained models in multilingual CEFR level assessment. Overall, \textsc{UniversalCEFR} aims to establish best practices in data distribution for language proficiency research by standardising dataset formats, and promoting their accessibility to the global research community.

\begin{tabular}{ll}
\raisebox{-0.3em}{\includegraphics[height=1em]{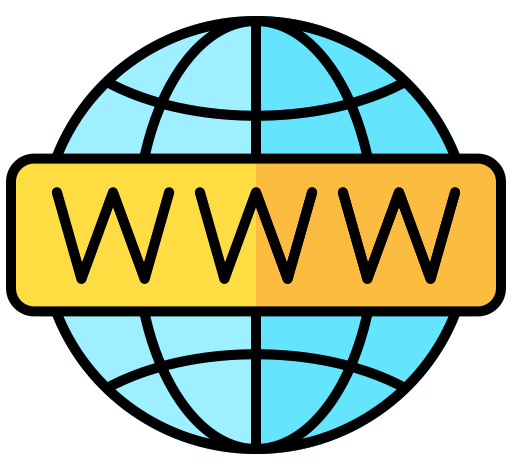}}  & \href{https://universalcefr.github.io}{\texttt{universalcefr.github.io}} \\
\raisebox{-0.3em}{\includegraphics[height=1em]{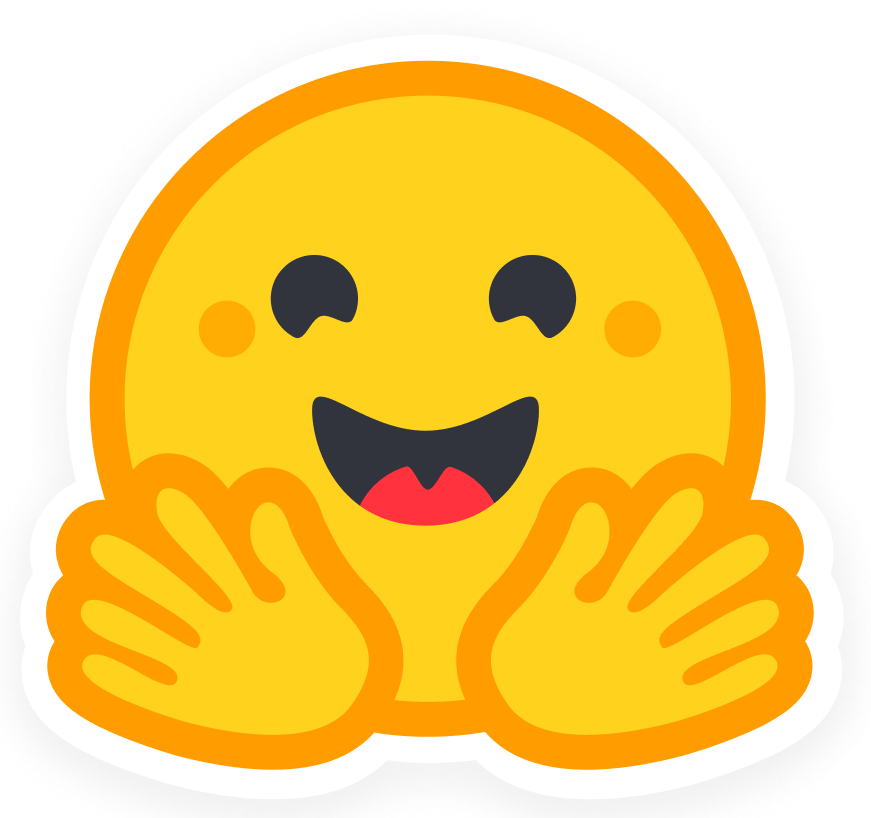}}  & \href{https://huggingface.co/UniversalCEFR}{\texttt{huggingface.co/UniversalCEFR}} \\
\raisebox{-0.3em}{\includegraphics[height=1em]{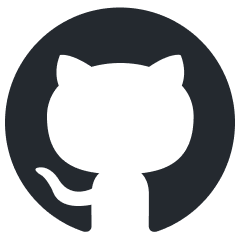}} & \href{https://github.com/UniversalCEFR}{\texttt{github.com/UniversalCEFR}} \\
\end{tabular}

\end{abstract}

\begin{figure}[!t]
    \centering
    \includegraphics[width=1.00\linewidth]{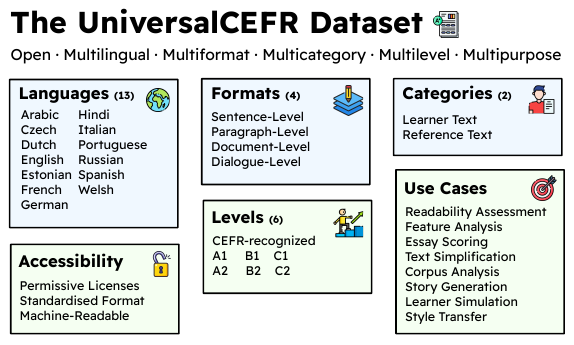}
    \caption{Overview of the contributions of the \textbf{\textsc{UniversalCEFR}} dataset, highlighting its \textbf{diverse structural coverage}—spanning language, format, category, and CEFR level—as well as its \textbf{accessibility and interoperability} for downstream tasks and use cases enabled by permissive licenses and standardized data formats.}
    \label{fig:universalcefr_dataset}
\end{figure}


\setlength{\tabcolsep}{3pt} 
\begin{table*}[t]
\centering
\small
\begin{tabular}{@{}lrrrllll@{}}
\toprule
\multicolumn{2}{l}{Resource} &
  \begin{tabular}[c]{@{}r@{}}\# Datasets\\ Indexed\end{tabular} &
  \begin{tabular}[c]{@{}r@{}}\# Languages\\Covered\end{tabular} &
  \begin{tabular}[c]{@{}l@{}}Data Types\end{tabular} &
  \begin{tabular}[c]{@{}l@{}}Data\\Accessibility\end{tabular} &
  \begin{tabular}[c]{@{}l@{}}Standard\\Format\end{tabular} &
  \begin{tabular}[c]{@{}l@{}}Geographic\\ Restrictions\end{tabular} \\ \midrule
\multicolumn{2}{l}{CEFRLex}                        & 7$^{\dag}$  & 6  & text               & unrestricted      & no  & none            \\
\multicolumn{2}{l}{Corpora @ UCLouvain}           & 31$^{\dag}$ & 9  & text, audio, video & request per corpus & no  & yes \\
\multicolumn{2}{l}{CLARIN L2 Learner Corpora}      & 75$^{\dag}$ & 34 & text, video        & request per corpus & no  & yes \\
\multicolumn{2}{l}{Learner Language (Språkbanken)} & 15$^{\dag}$ & 13 & text, audio        & request per corpus & no  & yes \\ \midrule
\multicolumn{2}{l}{\textbf{\textsc{UniversalCEFR}}}                  & \bf 26  & \bf 13 & \bf text               & \bf unrestricted    & \bf yes & \bf none            \\ \bottomrule
\end{tabular}
\caption{Comparison of existing language learning and language proficiency dataset collections with \textsc{UniversalCEFR}. $^{\dag}$ indicates that only a subset of the corresponding resource in that repository contains CEFR labels. Among the five repositories, \textsc{UniversalCEFR} is the only non-geo-locked and standardized collection, allowing seamless, unrestricted use for non-commercial research with proper attribution.}
\label{tab:resource_access_table}
\end{table*}

\section{Introduction}
Language proficiency research plays a central role in education, and often intersects with advances in linguistics and artificial intelligence (AI).
In natural language processing (NLP), language proficiency has been approached through well-established tasks such as automated readability assessment (ARA) and automated essay scoring (AES).
ARA focuses on determining whether a given text matches the expected reading skills of language learners according to their level, whereas AES evaluates the writing skills of the learners as reflected in a text they have written. 
In this paper, we combine these tasks under the more generic term of \textit{language proficiency assessment}, as it has varied practical applications in educational assessment and calibration of reading materials for learners \cite{xia-etal-2016-text,harsch2014general,figueras2012impact} as well as for various NLP tasks (see use cases in Figure \ref{fig:universalcefr_dataset}). A widely recognized standard for measuring second language (L2) proficiency is the Common European Framework of Reference for Languages (CEFR),\footnote{\url{https://www.coe.int/en/web/common-european-framework-reference-languages}} developed by the Council of Europe. 
CEFR offers a language-independent guide for evaluating learners' abilities in reading, writing, listening, and speaking.
It defines a six-level scale (A1, A2, B1, B2, C1, and C2) denoting increasing language competency \cite{north2014cefr,north2007cefr}. 

Recent advances in language proficiency assessment have moved from models relying on hand-crafted linguistic features to large language models (LLMs), which achieve high performance across diverse predictive and generative tasks through post-training techniques such as supervised fine-tuning \cite{devlin-etal-2019-bert,vaswani2017attention} or instruction tuning \cite{wei2022finetuned}. This form of task generalization enables complex linguistic pattern (e.g., features that make a text complex) modelling within unified frameworks for assessing language proficiency on standardized scales like CEFR. Moreover, they can also be extended to low-resource languages, potentially improving automatic assessment through techniques such as cross-lingual transfer \cite{he-li-2024-zero,imperial-kochmar-2023-automatic,imperial-kochmar-2023-basahacorpus,vajjala-rama-2018-experiments}. 

To fully leverage the potential of modern approaches for CEFR-level prediction, researchers require access to high-quality datasets with broad coverage across languages, proficiency levels, and text granularity. However, despite the long-standing use of CEFR in educational and NLP research, there are very limited standardized, machine-readable, and openly accessible collections of CEFR-annotated corpora, especially in terms of language coverage and granularity beyond sentence level \cite{naous-etal-2024-readme}. Moreover, most existing single-language resources are available in inconsistent or outdated formats (e.g., unprocessed text files, XML), which require extensive preprocessing and normalization. Finally, many datasets are restricted by copyright or licensing terms, limiting their accessibility for open research.

To this end, our work addresses the resource gap in CEFR-based language proficiency assessment research through the following contributions:
\begin{itemize}
    \item We introduce \textsc{UniversalCEFR}, a large-scale multilingual multidimensional open dataset composed of 505K CEFR-labeled texts across 13 languages, designed to advance multilingual research in language proficiency assessment.\vspace{-0.5em}
    \item We propose a data standardization pipeline and annotation template to homogenize available CEFR-labeled texts, enhancing their interoperability and accessibility for researchers across domains.\vspace{-0.5em}
    \item We provide a critical reflection of current practices in data sharing of language proficiency assessment resources and suggest pathways towards improvement using \textsc{UniversalCEFR} as a case study for a more open, standardized initiative for resource development.
\end{itemize}



\section{Background}
\label{sec:related_work}

\paragraph{Language Learning Databases and Resources.}
Language learning and language proficiency are research areas driven by the collection of two main types of data: reference-based data created by experts (e.g. reference reading materials) and learner-based data created by language learners (e.g. essays, conversations, and dialogue snippets). If a task requires it, such as in proficiency assessment, these corpora may undergo examination by language proficiency experts who will grade them based on a scale (e.g. CEFR). We list four community-recognized databanks and resource collections in the domain of language learning and proficiency assessment in Table~\ref{tab:resource_access_table}. CEFRLex is a collection of machine-readable multilingual lexicon-based datasets in 6 European languages. The Corpora Hub hosted by UCLouvain, the Learner Language from Språkbanken Text (SBX), and the L2 Learning Corpora hosted by the Common Language Resources and Technology Infrastructure (CLARIN) are all large collections of general multilingual and multimodal language learner datasets. Not all corpora in these databases are annotated with CEFR labels, and each corpus is associated with a publication detailing how they were collected and built and their specific purpose in language learning research. \\

\noindent \textbf{Access Restrictions and Data Privacy Regulations.} 
Despite the existence of L2 resource collections as listed in Table~\ref{tab:resource_access_table}, researchers cannot freely and openly use all datasets hosted in these repositories. CEFRLex,\footnote{\url{https://cental.uclouvain.be/cefrlex/}} Corpora @ UCLouvain,\footnote{\url{https://corpora.uclouvain.be/catalog/}} CLARIN,\footnote{\url{https://www.clarin.eu/resource-families/L2-corpora}} and Språkbanken Text\footnote{\url{https://spraakbanken.gu.se/en/resources/learner-language}} are hosted under European universities and institutions which means they are under the jurisdiction of EU Data Privacy Laws, particularly the General Data Protection Regulation (GDPR).\footnote{\url{https://gdpr-info.eu/}} Thus, learner texts from these collections, written based on personal interactions and containing Personally Identifiable Information (PII), can only be accessed through special legal coordination with the data maintainers. If access is granted, the licensee may also need to provide a proof of PII anonymization that produces a derivation distinct from the original dataset as done in \citet{jentoft-samuel-2023-nocola} for the ASK Corpus \cite{tenfjord-etal-2006-ask} containing L2 Norwegian CEFR-labeled texts and the International Corpus of Learner Finnish (ICLFI) \cite{ICLFI_en} containing L2 Finnish CEFR-labeled texts. Moreover, some datasets such as the SweLL Corpus \cite{volodina2024two,volodina2019swell,volodina-etal-2016-swell} from Språkbanken Text, composed of Swedish L2 texts with CEFR levels, are geographically licensed and can only be used by institutions within the EU and EEA region. As such, these datasets remain off-limits to any researcher outside of Europe.


\paragraph{CEFR Assessment and Standardization.}
The majority of research on automatic classification (or ranking) of texts based on the CEFR scale tends to focus on single-language model evaluations \cite{ribeiro-etal-2024-automatic,wilkens2024exploring,wilkens-etal-2023-tcfle,wilkens-etal-2018-sw4all,tack-etal-2017-human,volodina-etal-2016-swell,pilan2016readable,vajjala2014automatic,xia-etal-2016-text,yancey2021investigating,vasquez-rodriguez-etal-2022-benchmark}. This allows deeper investigation of language-specific nuances and intricacies connected to measuring text complexity. Meanwhile, other works have explored universal, language-agnostic features such as \citet{azpiazu-pera-2019-multiattentive,arhiliuc-etal-2020-language,caines-buttery-2020-reprolang,vajjala-rama-2018-experiments} where they used traditional word and PoS-ngram features to build a multi- and cross-lingual CEFR proficiency classifier for German, Czech, Italian, Spanish, and English, among others. \citet{he-li-2024-zero}, on the other hand, focused on cross-lingual automatic essay scoring anchored on the CEFR scale, covering six languages (Czech, English, German, Italian, Portuguese, and Spanish). 

In parallel with the rise of benchmarking studies for LLMs, similar efforts are growing in the CEFR-based language proficiency community. Two works in this direction include \citet{naous-etal-2024-readme}, which introduced ReadMe++, a multilingual, multidomain dataset for sentence-level readability assessment on a CEFR scale covering five languages, while the iRead4Skills Project by \citet{pintard2024iread4skills} released a collection of written texts in French, Portuguese, and Spanish across multiple genres and levels patterned to CEFR. Likewise, in data collection standardization, CLARIN released the Core Metadata Schema for Learner Corpora (LC-meta), which aims to provide a structured method with a specific emphasis on capturing metadata of collected learner texts, focusing on learner background, context, and individual differences \cite{paquot2024core}.


\section{The \textsc{UniversalCEFR} Dataset}
\label{sec:universal_cefr}

To support multilingual language proficiency research, we introduce \textsc{UniversalCEFR}, a large-scale initiative that curates and standardizes open human-annotated CEFR-labeled corpora. Unifying diverse resources under a consistent format enables reproducible and scalable research across linguistics, NLP, and education.
In this section, we outline the dataset's design principles, detail the data collection and standardization pipeline, provide key statistics, and present a linguistic feature analysis that supports downstream modelling. 

\subsection{Design Principles}
Our methodology was guided by three key design principles.

\paragraph{Openness and Accessibility.}
In building \textsc{UniversalCEFR}, we aim to demonstrate how data-driven research in language proficiency and assessment benefits from standardized, unified data formats. 
This enables portability and interoperability across domains with evolving data pipelines, such as language model pre-training in NLP. 
All corpora included in \textsc{UniversalCEFR} are publicly available for non-commercial research through permissive licenses (e.g. Creative Commons). 
However, significant effort was required to collate and standardize these datasets, highlighting the need for standardization and improved accessibility.

\paragraph{Multilinguality and Structure Diversity.}
Although CEFR originated in Europe, it has been increasingly adopted as a reference framework for language proficiency assessment worldwide. 
Accordingly, \textsc{UniversalCEFR} extends beyond European languages.
Its current version includes 13 languages, spanning high-resource (English, Spanish, French, German, Italian, Portuguese), mid-resource (Dutch, Russian, Arabic), and low-resource (Czech, Estonian, Hindi, Welsh) languages. 
It also captures structural diversity by annotating each corpus with its production category (learner or reference), granularity (sentence, paragraph, document, or discourse), and label coverage (standard CEFR or CEFR plus levels).

\paragraph{Global Collaboration.}
From its conceptualization and planning, the \textsc{UniversalCEFR} initiative involved close collaboration among 20 researchers in language proficiency assessment, NLP, and education from 13 institutions across nine countries (UK, Canada, USA, Germany, Sweden, UAE, Spain, Belgium, and Portugal).\footnote{As CEFR is a European framework, most active researchers in the field are based in Europe.} 
They all played a key role in defining the standardization protocol, designing evaluation experiments, and discussing future research directions.
These collaborative decisions are detailed in the following sections.

\subsection{Data Collection}
This section outlines the corpus selection criteria and the standardization methods used in \textsc{UniversalCEFR} for acquiring and consolidating a large and diverse collection of resources.

\paragraph{Corpora Selection.} 
The inclusion of datasets in \textsc{UniversalCEFR} is guided by three criteria: 

\begin{enumerate}
    \item \textbf{Public Accessibility:} Datasets must be available under a permissive license for non-commercial research (e.g., Creative Commons, CC-BY-NC), or be in the public domain and acquirable through direct download or via a request form for usage tracking.

    \item \textbf{Gold-Standard CEFR Labels}: Datasets must include CEFR annotations produced or validated by domain experts, such as language teachers or proficiency researchers, particularly in the case of learner texts.

    \item \textbf{Human Authorship:} All texts must be written by humans to ensure suitability for research involving creative, multilingual, multi-level, and multi-genre content. As of this writing, \textsc{UniversalCEFR} does not include machine-generated texts. 
\end{enumerate}


The full list of consolidated corpora that meet all three \textsc{UniversalCEFR} inclusion criteria is provided in Table~\ref{tab:universalcefr_full_table} in the Appendix.

\paragraph{Standardization Process.} 
To ensure interoperability, transformation, and machine readability, we standardized the collected datasets by preprocessing their varied source formats into a unified structure.
We adopted JSON as the per-instance format and defined eight metadata fields considered essential for each CEFR-labeled text. 
These fields include the source dataset, language, granularity (document, paragraph, sentence, discourse), production category (learner or reference), and license. 
Full descriptions and predetermined values used for each field are provided in Table~\ref{tab:json_fields}. 
The final standardized dataset is available from HuggingFace Dataset repository.\footnote{\url{https://huggingface.co/UniversalCEFR/datasets}}
A key challenge was the lack of a unified format across the language proficiency community. 
Source corpora came in various formats, including plain text (e.g., csv, tsv, txt), spreadsheets (e.g., XLSX, XLS), markup (e.g., XML), and PDFs requiring manual extraction.
This challenge further motivates the need for unified data aggregation initiatives that \textsc{UniversalCEFR} aims to help establish.


\setlength{\tabcolsep}{10pt} 
\begin{table}[!t]
\small
\centering
\begin{tabular}{@{}llr@{}}
\toprule
\multicolumn{2}{l}{\textsc{UniversalCEFR}} & {\# of Instances} \\
\midrule
\multicolumn{2}{l}{\textsc{  - full*}}                 & 505,807                   \\
\multicolumn{2}{l}{\textit{    (out-of-scope instances)}}                 & 11,316                   \\
\hdashline 
\multicolumn{2}{l}{\textsc{  - full}}        & 494,491                   \\
\multicolumn{2}{l}{\textsc{  - train}}        & 435,919                   \\
\multicolumn{2}{l}{\textsc{  - dev}}          & 54,107                    \\
\multicolumn{2}{l}{\textsc{  - test}}         & 4,465                     \\
\bottomrule
\end{tabular}
\caption{Data splits for {\textsc{UniversalCEFR}}.  \textsc{full*} denotes all instances, including those with CEFR labels that we currently do not recognize for the task (e.g., NA, A+, B). These were excluded from the \textsc{train}, \textsc{dev}, and \textsc{test} sets used in our experiments.}
\label{tab:universalcefr_split}
\end{table}

\subsection{Dataset Statistics}
The final \textsc{UniversalCEFR} collection comprises \textbf{505,807 CEFR-labeled texts} across \textbf{13 languages} and \textbf{4 scripts} (Latin, Arabic, Devanagari, and Cyrillic). 
Tables~\ref{tab:universalcefr_split} and ~\ref{tab:universalcefr_main_count_grade_coverage} show the overall dataset size, its splits and breakdown per CEFR level per language. We identified 11,316 instances with invalid or out-of-scope labels (e.g., NA, A+, B) outside the six recognized CEFR labels (A1–C2) and duplicates, which were removed before splitting \textsc{UniversalCEFR} into \textsc{train}, \textsc{dev}, and \textsc{test}. 
For the \textsc{test}, we set a cap of 200 instances per language and per granularity level. 
Additional dataset statistics can be found in Appendix~\ref{app:full_data_statistics}.

\setlength{\tabcolsep}{3pt} 
\begin{table}[htbp]
\centering
\footnotesize
\begin{tabular}{@{}lrrrrrr@{}}
\toprule
{\textsc{Lang}}  & {\textsc{A1}}     & {\textsc{A2}}     & {\textsc{B1}}     & {\textsc{B2}}    & {\textsc{C1}}    & {\textsc{C2}}   \\ \midrule
EN & 192,596 & 132,614 & 66,425 & 23,266 & 8,004 & 795 \\
ES & 8,282    & 8,648    & 6,835  & 5,061  & 3,224 & 0   \\
DE & 319     & 15,970  & 15,630 & 474    & 130   & 426 \\
NL & 51      & 216     & 782    & 738    & 219   & 85  \\
CS & 1       & 188     & 165    & 81     & 4     & 0   \\
IT & 29      & 381     & 394    & 2      & 0     & 0   \\
FR & 151     & 390     & 575    & 478    & 293   & 126 \\
ET & 0       & 395     & 588    & 407    & 307   & 0   \\
PT & 314     & 325     & 367    & 233    & 112   & 72  \\
AR & 81      & 259     & 625    & 645    & 361   & 183 \\
HI & 263     & 283     & 286    & 263    & 222   & 174 \\
RU & 402     & 293     & 409    & 326    & 237   & 91  \\
CY & 764     & 608     & 0      & 0      & 0     & 0   \\ \midrule
\textbf{Total} & \textbf{203,253} & \textbf{160,570} & \textbf{93,081} & \textbf{31,974} & \textbf{13,113} & \textbf{1,952} \\ \bottomrule
\end{tabular}
\caption{Data statistics of \textbf{\textsc{UniversalCEFR-full}} in terms of recognized CEFR levels (A1, A2, B1, B2, C1, C2) across the 13 target languages.}
\label{tab:universalcefr_main_count_grade_coverage}
\end{table}

\section{Linguistic Feature Analysis}
\label{sec:linguistic_feature_analysis}
We aim to examine how well a broad set of linguistic features aligns with CEFR proficiency levels across languages in \textsc{UniversalCEFR}.
We extracted a set of \textbf{100 linguistic features}, grouped into morphosyntactic (62), syntactic (18), length-based (11), lexical (4), readability (2), psycholinguistic (2), and discourse (1) categories. 
A complete and detailed list is available in Appendix \ref{app:appendix-features}. 

\subsection{Correlation Across All Languages} Considering the absolute Spearman correlation between the features and the CEFR level (selecting values with $p < 0.05$ and $\rho > 0.3$ on average across all languages), the strongest associations were found in length-based measures, such as characters per sentence and syllables per sentence. Several grammatical complexity features, including parse tree height and phrase length, showed moderate correlations. Readability indices (FKGL and Flesch Reading Ease) also displayed moderate correlations in the expected direction. Psycholinguistic features, such as concreteness and imageability, were negatively correlated with proficiency, indicating a shift toward more abstract language at higher levels. Finally, morphosyntactic features regarding voice, tense, and number showed moderate but consistent correlations, supporting their relevance in reflecting syntactic development.

\definecolor{myblue}{RGB}{59,76,192}
\definecolor{myred}{RGB}{180,4,38}
\begin{figure}[!t]
    \centering
    \includegraphics[width=1.00\linewidth]{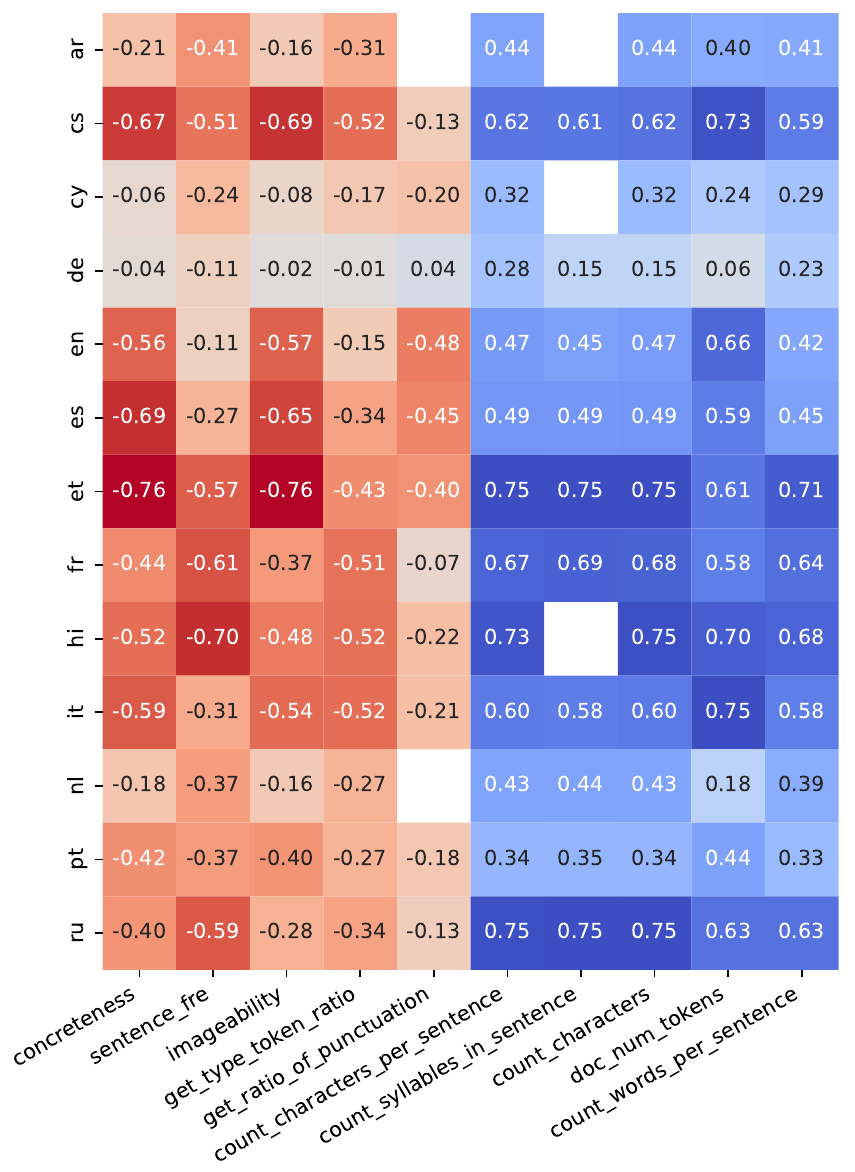}
    \caption{Highly correlated linguistic features occurring in at least three languages. \textcolor{myblue}{\bf Blue features} lean towards positive correlation, while \textcolor{myred}{\bf red features} denote negative correlation. For brevity, these top features are those lying at the extreme ends of the correlation spectrum.}
    \label{fig:heatmap_linguistic}
\end{figure}

\subsection{Correlation By CEFR Level} To assess the consistency of feature relevance across languages, we examined the number of features with significant correlations ($p < 0.05$) with CEFR levels per language as visualized in Figure~\ref{fig:heatmap_linguistic}. The results revealed notable variations. Languages such as Czech (\textsc{cs}), Estonian (\textsc{et}), and Italian (\textsc{it}) showed a high number of relevant features, suggesting strong alignment between the selected linguistic features and CEFR progression in these languages. English (\textsc{en}), Spanish (\textsc{es}), French (\textsc{fr}), Hindi (\textsc{hi}), and Russian (\textsc{ru}) showed moderate coverage, with a reasonable number of features exceeding the 0.3 correlation threshold. In contrast, Arabic (\textsc{ar}), Dutch (\textsc{nl}), and Portuguese (\textsc{pt}) exhibited weak coverage, while Welsh (\textsc{cy}) and German (\textsc{de}) had very few or no features with relevant correlations, indicating a limited match between the current feature set and CEFR levels for those languages. Furthermore, a few features are only relevant for a few languages, e.g., the translative case for only Estonian, negative verb polarity for only Czech, or genitive case for only Czech, Estonian, and Russian. This variability highlights the influence of language-specific properties on the effectiveness of general feature-based models for proficiency prediction.

\setlength{\tabcolsep}{3.5pt} 
\renewcommand{\arraystretch}{1.2}
\begin{table*}[!t]
\small
\centering
\begin{tabular}{@{}lcccccccccccccc@{}}
\toprule
\textsc{Model \& Setup} &
  \textsc{EN} &
  \textsc{ES} &
  \textsc{DE} &
  \textsc{NL} &
  \textsc{CS} &
  \textsc{IT} &
  \textsc{FR} &
  \textsc{ET} &
  \textsc{PT} &
  \textsc{AR} &
  \textsc{HI} &
  \textsc{RU} &
  \textsc{CY} & 
  {Avg} \\ \midrule

\rowcolor{gray!10}
\multicolumn{1}{@{}l}{\textbf{\textsc{Baseline}}}
  &   &   &   &   &   &   &   &   &   &   &   &   &   &   \\ 
  \hspace{1em}\textnormal{\indent\textsc{Most Frequent Class}}
  & 7.39            & 18.1            & 26.8  & 21.4            & 23.8            & 35.5            & 16.3  & 15.9            & 10.0            & 23.3            & 7.28  & 10.7  & 33.4            & 19.3            \\

\rowcolor{gray!20}
\multicolumn{1}{@{}l}{\textbf{\textsc{Gemma1-7B (English)}}}
  &   &   &   &   &   &   &   &   &   &   &   &   &   &   \\ 
  \hspace{1em}\textnormal{\indent\textsc{Base}}
  & \underline{21.8}            & 26.0            & \underline{40.6}  & 32.1            & 44.0            & 57.3            & \underline{32.2}  & 39.0            & 14.0            & 28.9            & \underline{25.0}  & \underline{34.8}  & 48.7            & 34.2            \\
  \hspace{1em}\textnormal{\indent\textsc{En-Read}}
  & 20.5            & 28.3            & 31.0            & 23.5            & 53.6            & 41.0            & 22.7            & 24.9            & \underline{27.2}  & 29.5            & 8.4             & 18.0            & \underline{55.7}  & 29.6            \\
  \hspace{1em}\textnormal{\indent\textsc{En-Write}}
  & {19.8}  & 24.5            & 34.5            & 29.3            & 51.9            & 57.7            & 27.7            & 42.7            & 22.2            & 20.8            & 14.0            & 27.6            & 52.1            & 32.9            \\
  \hspace{1em}\textnormal{\indent\textsc{Lang-Read}}
  & {20.5}            & 29.3            & 35.1            & \underline{37.8}  & \underline{55.3}  & 48.0            & 27.1            & 44.6            & 20.2            & 32.2            & 12.8            & 26.2            & 52.8            & 34.0            \\
  \hspace{1em}\textnormal{\indent\textsc{Lang-Write}}
  & {19.8}  & \underline{29.8}  & 32.6            & 34.0            & 49.9            & \underline{61.7}  & 26.3            & \underline{46.3}  & 21.2            & \underline{36.7}  & 12.7            & 26.9            & 53.6            & \textbf{34.7}   \\
\midrule

\rowcolor{gray!20}
\multicolumn{1}{@{}l}{\textbf{\textsc{Gemma3-12B (Multi)}}}
  &   &   &   &   &   &   &   &   &   &   &   &   &   &   \\ 
  \hspace{1em}\textnormal{\indent\textsc{Base}}
  & \underline{28.8}            & \underline{35.0}  & 42.2            & \underline{47.0}  & 42.6            & 65.2            & 38.1            & 39.5            & 24.6            & 41.8            & \underline{28.7}  & 29.7            & 40.9            & 38.8            \\
  \hspace{1em}\textnormal{\indent\textsc{En-Read}}
  & 19.3            & 25.5            & 35.8            & 25.5            & 18.5            & 22.9            & 29.3            & 26.0            & 9.8             & 33.3            & 14.8            & 21.2            & 20.5            & 23.3            \\
  \hspace{1em}\textnormal{\indent\textsc{En-Write}}
  & 26.6  & 36.7            & \underline{46.4}  & 46.7            & 50.1            & \underline{77.4}  & \underline{40.5}  & \underline{43.8}  & \underline{27.3}  & \underline{48.6}  & 24.0            & \underline{37.4}  & 52.4            & \textbf{43.2}   \\
  \hspace{1em}\textnormal{\indent\textsc{Lang-Read}}
  & 19.3            & 28.1            & 35.2            & 37.6            & 50.9            & 64.8            & 35.0            & 30.4            & 26.1            & 29.5            & 20.5            & 32.5            & \underline{61.6}  & 36.3            \\
  \hspace{1em}\textnormal{\indent\textsc{Lang-Write}}
  & 26.6  & 33.2            & 38.3            & 39.6            & \underline{55.0}  & 76.4            & 37.7            & 42.4            & 25.4            & 38.0            & 24.6            & 31.5            & 53.7            & 40.2            \\
\midrule

\rowcolor{gray!20}
\multicolumn{1}{@{}l}{\textbf{\textsc{EuroLLM-9B (Multi)}}}
  &   &   &   &   &   &   &   &   &   &   &   &   &   &   \\ 
  \hspace{1em}\textnormal{\indent\textsc{Base}}
  & 18.6            & 25.4            & 28.0            & 29.1            & 25.0            & 39.9            & 25.9            & 32.0            & 16.4            & 34.3            & 12.7            & 15.1            & 14.4            & 24.4            \\
  \hspace{1em}\textnormal{\indent\textsc{En-Read}}
  & \underline{23.1}  & 26.9            & \underline{38.1}  & 30.2            & \underline{33.3}  & \underline{41.9}  & 24.5            & \underline{33.6}  & 19.9            & 33.8            & 18.0            & \underline{21.8}  & \underline{26.4}  & \textbf{28.6}   \\
  \hspace{1em}\textnormal{\indent\textsc{En-Write}}
  & 21.5            & 26.2            & 29.8            & \underline{32.0}  & 32.4            & 33.1            & 26.8            & 32.8            & \underline{21.1}  & 31.8            & 17.7            & 17.5            & 24.5            & 26.7            \\
  \hspace{1em}\textnormal{\indent\textsc{Lang-Read}}
  & \underline{23.1}  & 27.0            & 32.7            & 31.8            & 29.8            & 32.9            & \underline{28.3}  & 28.6            & 16.8            & 32.4            & 14.3            & 16.2            & 17.3            & 25.5            \\
  \hspace{1em}\textnormal{\indent\textsc{Lang-Write}}
  & 21.5            & \underline{28.5}  & 35.1            & 30.1            & 30.8            & 30.6            & 27.6            & 29.9            & 16.5            & \underline{35.2}  & \underline{21.0}  & 16.1            & 8.80             & 25.5            \\ \midrule

\rowcolor{gray!20}
\multicolumn{3}{@{}l}{\textbf{\textsc{fine-tuned Models}}}  &   &   &   &   &   &   &   &   &   &   &   &   \\
\hspace{1em}\textnormal{\indent\textsc{ModernBERT (English)}}
 & \underline{75.8} & 71.8 & 72.1 & 54.2 & 66.9 & 82.7 & 47.2 & 88.3 & \underline{33.5} & 30.8 & 51.6 & 48.9 & 73.2 & 61.3 \\

\hspace{1em}\textnormal{\indent\textsc{EuroBERT (Multi)}}
  & 74.6            & \underline{72.0}            & 70.6            & 53.2            & 63.9            & 79.7            & 42.0            & 86.6            & 32.1            & 35.4            & 44.7            & 45.9            & \underline{79.9}            & 60.0            \\
\hspace{1em}\textnormal{\indent\textsc{XLM-R (Multi)}}
  & 75.5            & 69.6            & \underline{73.2}            & \underline{59.0}            & \underline{68.8}            & \underline{83.2}            & \underline{51.6}            & \underline{88.8}            & 29.2            & \underline{43.0}            & \underline{52.8}            & \underline{49.6}            & 72.6            & \textbf{62.8}            \\
  \midrule

\rowcolor{gray!20}
\multicolumn{3}{@{}l}{\textbf{\textsc{Feature-Based Models}}}  &   &   &   &   &   &   &   &   &   &   &   &   \\
\hspace{1em}\textnormal{\indent\textsc{RandForest (TopFeats)}}
  & 62.0 & 57.6 & 64.9 & \underline{54.5} & \underline{69.5} & 79.9 & \underline{44.1} & \underline{84.2} & \underline{27.8} & \underline{43.8} & 44.1 & 47.2 & 72.9
 & 57.9            \\
\hspace{1em}\textnormal{\indent\textsc{RandForest (AllFeats)}}
  & \underline{63.4} & \underline{60.6} & \underline{65.4} & 53.0 & 69.2 & 79.3 & 41.4 & \underline{84.2} & 26.4 & 42.8 & 46.8 & \underline{47.8} & \underline{78.2} & \textbf{58.3}            \\

\hspace{1em}\textnormal{\indent\textsc{LogRegr (AllFeats)}} & 32.1 & 28.2 & 50.9 & 47.1 & 62.9 & 81.9 & 41.7 & 67.5 & 23.1 & 34.1 & 47.8 & 41.1 & 63.8& 47.9 \\
\hspace{1em}\textnormal{\indent\textsc{LogRegr (TopFeats)}}
  & 30.4 & 29.7 & 52.5 & 44.1 & 62.7 & \underline{82.7} & 40.3 & 67.5 & 22.7 & 33.5 & \underline{48.4} & 41.1 & 59.2 & 47.3 \\
  
\bottomrule
\end{tabular}

\caption{Full weighted F1 performance results from the multilingual and English-centric model evaluation experiments using three setups (feature-based, fine-tuning, and prompting) and using \textsc{UniversalCEFR-test} split across the 13 languages. \textbf{Boldfaced} values indicate the highest scores overall per model setup, while \underline{underlined} values highlight the highest scores for each model setup within each language.} 
\label{tab:main_results_grouped}

\end{table*}

\section{CEFR Level Classification}
\label{sec:cefr_level_classification}

Given the availability of gold-standard CEFR labels and the linguistic diversity of the \textsc{UniversalCEFR} dataset, we define our primary experimental task as \textbf{multiclass, multilingual CEFR level classification}. 
The goal is to predict one of the six CEFR levels (A1–C2) for a given text instance in any of the 13 supported languages.
We evaluate three modeling paradigms: feature-based classification, fine-tuning of multilingual pre-trained models, and prompting LLMs.




\subsection{Feature-Based Models}
We evaluated two widely-used classification models from Scikit-Learn~\cite{scikit-learn}: \textbf{Random Forest} (\textsc{RandForest}) and \textbf{Logistic Regression} (\textsc{LogRegr}).
Both models were trained on the linguistic features described in Section~\ref{sec:linguistic_feature_analysis}, using Scikit-Learn’s default hyperparameter settings.
We experimented with two feature configurations: one using all 100 features (\textsc{AllFeats}) and another using an automatically selected subset of top-performing features across all languages (\textsc{TopFeats}).  Appendices \ref{app:appendix-allfeatures} and \ref{app:appendix-topfeatures} detail the linguistic feature information for both setups.

\subsection{Fine-tuned Models}
We used three BERT-based models with varying degrees of multilingual coverage: \textbf{ModernBERT} \citep{modernbert}, a monolingual English model with 395M parameters; \textbf{EuroBERT} \citep{EuroBERT}, a multilingual model trained on 15 diverse European and non-European languages, with 210M parameters; and \textbf{XLM-R} \citep{conneau-etal-2020-unsupervised}, a massively multilingual model supporting 100 languages, with 279M parameters. 
Each model was fine-tuned for three epochs, with the best checkpoint selected based on the highest weighted F1 score on the validation set. Additional details  can be found in Appendix Table~\ref{tab:fine-tuning_hyperparams}.

\subsection{Descriptor-Based Prompting}
We evaluated three instruction-tuned models: \textbf{Gemma 1}~\citep{team2024gemma}, an English-centric model with 7B parameters; \textbf{Gemma 3}~\citep{team2025gemma}, a multilingual model trained on 140+ global languages with 12B parameters; and \textbf{EuroLLM}~\citep{martins2024eurollm}, a multilingual model trained on 15 European-centric languages with 9B parameters. 
We explored five prompting strategies, ranging from no context to setups using CEFR level descriptors for reading comprehension and written production, either in English or in specific languages. 
The prompt configurations are as follows:

{\small
\begin{itemize}
    \item \textbf{\textsc{Base}}. Generic prompting with no CEFR level descriptors as context.\vspace{-0.5em}
    
    \item \textbf{\textsc{En-Read}}. CEFR level descriptors for reading comprehension in English used as  context.\vspace{-0.5em}
    
    \item \textbf{\textsc{En-Write}}. CEFR level descriptors for written production in English used as context.\vspace{-0.5em}
    
    \item \textbf{\textsc{Lang-Read}}. CEFR level descriptors for reading comprehension, translated to the target language being assessed used as context.\vspace{-0.5em}
    
    \item \textbf{\textsc{Lang-Write}}. CEFR level descriptors for written production, translated to the target language being assessed used as context.
\end{itemize}}

All CEFR descriptors were retrieved from the official CEFR website.
Prompt templates and hyperparameter values for each setup are detailed in Table~\ref{tab:promtping_hyperparams} and Appendix \ref{app:prompt_templates}.

\subsection{Evaluation Metrics}
We use \textbf{weighted F1} as the primary evaluation metric across all experiments. This accounts for the class imbalance in CEFR level distribution and granularity across language subsets in \textsc{UniversalCEFR-test}. Using accuracy in the experiments would produce misleading performance in favor of any majority class.

\section{Results}
\label{sec:results}

\subsection{Model-Based Performance Comparison}
Table~\ref{tab:main_results_grouped} shows that, in terms of overall average performance across languages, the fine-tuned setup with ModernBERT, EuroBERT, and XLM-R achieved the highest weighted F1 score range ($\approx$60\%-62.8\%) outperforming feature-based models ($\approx$47\%-58\%) and prompting ($\approx$23\%-43\%). 
Among the LLM-based approaches---prompting and fine-tuning---models trained on broader multilingual corpora generally performed better. 
For instance, XLM-R, which supports 100 languages, was the top performer, followed by EuroBERT (15 languages) and ModernBERT (English-only). 
A similar trend was observed in prompting: Gemma 3, trained on 140+ languages, outperformed EuroLLM (15 languages) and the English-centric Gemma 1, achieving the best prompting score of 43.2.
These findings are consistent with previous work \citep{naous-etal-2024-readme,shardlow-etal-2024-bea,colla-etal-2023-elicode,yuan-strohmaier-2021-cambridge}, reinforcing the usefulness of multilingual models for language proficiency assessment tasks.
One limitation of our experimental setup, however, is that we did not include language-specific pre-trained models for languages other than English, which may have further improved performance for low- and mid-resource languages.

\setlength{\tabcolsep}{5pt}
\begin{table}[t]
\centering
\small 
\begin{tabular}{@{}lrrrr@{}} \toprule
{\textsc{Model}} & {\textsc{Sent}} & {\textsc{Para}} & {\textsc{Doc}} & {\textsc{All}}\\ \midrule

\textsc{Gemma1} & 19.41 & \textbf{42.74} & 30.81 & 33.63\\

\textsc{Gemma3} &38.71 & \textbf{43.12}& 39.62 &42.33\\

\textsc{XLM-R} & 62.67& 66.38 &\textbf{71.12} &65.92\\

\textsc{RandForest-All} & 56.88 & 62.77&\textbf{64.58}&61.38 \\

\textsc{RandForest-Top} & 53.89& 62.98& \textbf{64.94}&60.50\\  

\bottomrule
\end{tabular}
\caption{Weighted F1 scores for top-performing unique model evaluation setups across granularities available for all languages.}
\label{tab:granular_results}
\end{table}

\subsection{Granularity-Level Comparison}
\label{subsec:granularity}
Table~\ref{tab:granular_results} highlights clear performance differences across text granularities (sentence, paragraph, and document) for all models, but more prominently for the Gemma models under prompting. 
Gemma 1, in particular, tends to over-predict lower CEFR levels (A1–B1) on sentence-level data, whereas its predictions on document-level subsets are more evenly distributed and better aligned with ground truth distributions.
This suggests that prompt-based methods may require longer texts to make more accurate predictions, unlike models trained or fine-tuned on the respective datasets.
Other models, such as XLM-R and Random Forest, show better results on document ($\approx$64\%-71\%) and paragraph-level data ($\approx$62\%-66\%) than sentence-level data ($\approx$53\%-62\%), which was shown to be a more difficult task in previous work on readability \cite{dell2011read,vajjala2014assessing}. 
Regarding language-specific differences, among English, German, and Welsh, the best performance is seen with the paragraph-level dataset for English, the document-level dataset for German, and the sentence-level dataset for Welsh and French with the fine-tuned XLM-R model. 
Similar variations can be observed for other languages with more than one level of granularity (see Table~\ref{tab:granular_lang_perf}).
No single granularity or model shows consistently better performance across all tested languages. These results are likely due to the distribution of excerpts across granularity levels in each language (see Table \ref{tab:universalcefr_main_count_level} in Appendix~\ref{app:full_data_statistics}).

\subsection{Learner-Reference Comparison}
Four languages in \textsc{UniversalCEFR} contain both learner and reference texts: Arabic, German, English, and Spanish. 
Table~\ref{tab:learn-ref-diff} reports the average weighted F1 performance difference between the two categories across the four languages.
For German, performance is comparable between learner and reference texts ($\approx$71–74\%). In contrast, English and Spanish show higher performance on learner texts (83\% and 98\%) than on reference texts (58\% and 42\%, respectively).
Arabic displays the opposite trend: results on reference texts (54\%) are much higher than those of learner texts, where the best results were obtained by Gemma 3 (41\%). 
One possible explanation is that Gemma 3 may have been exposed to more Arabic content in its pre- and post-training phases.


\begin{table}[t]
\centering
\small
\begin{tabular}{@{}lrr@{}} \toprule 
{\textsc{Language}} & {\textsc{Learner}} & {\textsc{Reference}} \\ \midrule
\textsc{AR} & 41.92$^\dag$ & 54.69 \\
\textsc{DE} & 71.14 & \textbf{74.39} \\
\textsc{EN} & 83.41 & 58.24 \\
\textsc{ES} & \textbf{97.99} & 42.72 \\ \bottomrule
\end{tabular}
\caption{Average performances of the best models on learner text versus reference text across languages.$^\dag$ indicates performance with Gemma 3, and the rest refer to performance of the XLM-R model. Only these four languages have both learner and reference texts.}
\label{tab:learn-ref-diff}
\end{table}


\section{Discussion}
\label{sec:discussion}
We discuss potential pathways through which  \textsc{UniversalCEFR} can serve as a model, and offer key considerations for advancing data accessibility in language proficiency research.

\paragraph{Critical Reflections of Current Practices.}
The multiregional and multidisciplinary effort behind \textsc{UniversalCEFR} exposed significant inconsistencies and critical gaps in building CEFR-labeled language proficiency assessment corpora. 
Upon examination of annotation practices, \textit{there appears to be no standard method for conducting expert annotations, including inconsistent use of inter-annotator agreement metrics and unclear guidelines on the number of annotators required to achieve reliable agreement}.
This is reflected in the \textsc{UniversalCEFR} dataset itself, where nearly half of the corpora lack information on the annotators involved and their agreement scores. 
We posit that this may be due to diverse judgments of what constitutes high-quality data that does not require further human annotations.

In terms of language coverage, \textsc{UniversalCEFR} includes nine ({\textsc{EN, ES, DE, NL, CS, IT, FR, ET, PT}}) of the 24 recognized European languages. 
As a result, researchers working on these nine languages now have access to open, standardized data for CEFR-based language proficiency assessment. 
The remaining 15 languages represent valuable opportunities for future expansion through collaborative efforts.
While our open data and standardization initiative is a step towards addressing current challenges in interoperability and accessibility of resources, similar parallel efforts are needed in areas such as annotation and evaluation practices to ensure sustained progress in the language proficiency assessment community.


\paragraph{Need for Pro-Research Data Sharing Policies.} 
As generative AI, particularly LLMs, becomes more ubiquitous, 
organizations that create valuable data for language proficiency assessment, such as publishers, educational institutions, and media outlets, are growing more cautious about how their resources are used.
A major concern is the risk of data being used to train proprietary generative models, especially when such models are only accessible via commercial APIs that require transferring evaluation corpora to external servers.
An example is the TCFLE-8 corpus \cite{wilkens-etal-2023-tcfle} containing CEFR-labeled essays hosted by France Education International. 
Researchers seeking access to this dataset must explicitly specify that the resource will not be processed through commercial APIs to prevent potential data harvesting. 
To address these concerns, we believe \textit{the community needs to agree on a unified pro-research data sharing policy with clear usage guidelines} for academic, non-commercial studies that require analysis of protected data with generative AI models without training on them.

\paragraph{Linguistic Features and Fine-tuning Still Matter.}
While recent advances in LLMs keep transforming NLP research, \textit{our multilingual and multidimensional experiments in Section~\ref{sec:results} reaffirm the continued value of linguistic features for traditional ML classifiers and fine-tuning pre-trained models} in language proficiency assessment. 
We observe common patterns where higher distribution and instance count lead to better results using these two setups (see performances on Spanish, English, and German subsets in Table~\ref{tab:main_results_grouped}) over prompting with CEFR descriptors. 
Moreover, using linguistic features in language proficiency assessment allows deeper analysis of language interactions with variables such as complexity, as seen in Appendix~\ref{app:language_specific_analysis}. 
Given these insights, we encourage further efforts in the expansion of existing but low-resource language datasets with CEFR labels, as well as the exploration of features to better model morphologically-rich languages (e.g., Estonian and Portuguese). Together, these recommendations bridge current observed model failures to practical approaches in improving multilingual CEFR proficiency assessment.




\section{Conclusion and Future Directions}
In this work, we introduced \textsc{UniversalCEFR}, a large-scale, open, multilingual, multidimensional dataset comprising 505,807 CEFR-annotated texts across 13 languages developed through global collaboration. 
Our findings from diverse model experiments with CEFR level prediction provide strong support for the utility of linguistic features and fine-tuning multilingual models in language proficiency assessment. 
Similarly, our critical analysis of the current data and resource-building practices emphasized the need for similar initiatives from the community, and pro-research data sharing policies in the advent of generative AI to remove barriers to accessibility without compromising data privacy and intellectual property.

Beyond its data and technical contributions, \textsc{UniversalCEFR} also carries broader sociolinguistic significance. \textsc{UniversalCEFR} addresses the growing linguistic inequality in modern AI development by focusing on underrepresented languages alongside English. We hope this initiative can lead to more responsible AI development that actively resists the growing linguistic centralization around English in global AI research---a modern \textit{Matthew effect} \cite{merton1988matthew}---where well-resourced languages receive disproportionate technological attention while smaller languages (like Czech or Welsh) are left behind \cite{masciolini2025towards}. The \textsc{UniversalCEFR} is a strong step towards mitigating the Matthew effect in language proficiency assessment research.


\section*{Limitations}
We discuss several limitations of our work on \textsc{UniversalCEFR} and how researchers can consider these directions to develop the resource further.

\paragraph{Natural Data Disparity in Experiments.}
From the statistics presented in Tables~\ref{tab:universalcefr_main_count_grade_coverage} and \ref{tab:universalcefr_main_count_level} for \textsc{UniversalCEFR}, it is expected that not all languages have the exact same distribution of data across dimensions, including formats (sentence-, paragraph-, document-, and dialogue-level) and category (reference and learner texts). Hence, our main experiments in Table~\ref{tab:main_results_grouped} combined these variables to provide a unified performance comparison. We note that while this offers a broad overview of the three evaluation paradigms (prompting, fine-tuning, and linguistic features) across languages, future work should include dedicated modeling and evaluation by text category, which may warrant a more focused, in-depth study we assign for future work.


\paragraph{Language Availability and Dependency.} 
Due to the nature of \textsc{UniversalCEFR} being a standardized collection of open-sourced, publicly accessible CEFR data, its growth depends heavily on how the community will move forward and continuously release artifacts, including CEFR-annotated corpora for reproducibility and wider access for research purposes. We also acknowledge the efforts of researchers who work on multi-framework adoption, where CEFR descriptors and bands are overlapped with languages not within Europe (such as Hindi \cite{naous-etal-2024-readme} and Arabic \cite{habash-palfreyman-2022-zaebuc}), and continue to open-source the annotated data.

\paragraph{Modalities Beyond Texts.}
The current data collection scope of \textsc{UniversalCEFR} and the insights presented in this work only cover CEFR-based texts for now, specifically for reading and writing specifications. Multimodal data, such as audio and video recordings of learners associated with CEFR specifications for listening and speaking, are not yet covered. Naturally, these datasets are even more challenging to acquire and open-source, especially if they contain materials from or are created by learners under legal age and if they contain personal information.

\paragraph{Beyond Typical Benchmarking}
The rigor of analysis in this paper is not meant to be treated as a typical benchmark study, similar to recent trends in NLP papers, where the goal is to evaluate as many LLMs as possible. In this paper, we provide deeper insights into language complexities and intricacies that affect model performance in CEFR level classification across various dimensions of language, granularity, and format. Thus, within our compute budget, we carefully handpicked state-of-the-art LLMs that are worth exploring based on their properties (e.g., English-centric against massively multilingual, or linguistic features against fine-tuning and prompting). We leave the evaluation on larger, more advanced LLMs, as well as explorations in other directions to improve CEFR level classification, such as the use of high-quality synthetic datasets, for future work.


\section*{Ethics Statement}
As mentioned throughout this paper, all the datasets we collected for \textsc{UniversalCEFR} based on our criteria presented in Section~\ref{sec:universal_cefr} are already publicly accessible with permissive licenses, and can be used for non-commercial research purposes. While there are three corpora from \textsc{UniversalCEFR}---namely APA-LHA, DEplain, EFCAMDAT---that require users to fill a short form and agree to terms, we still classified them as publicly accessible due to the quick response to access approval. 

In the context of the EU AI Act, the use of AI systems for educational purposes, especially those that are intended to \textit{"to evaluate learning outcomes, including when those outcomes are used to steer the learning process of natural persons in educational and vocational training institutions at all levels"} \cite{aiact2024}, is classified under \textit{high risk}. Thus, AI systems that will be released in the market with these goals are required to comply with obligations for high-risk systems, including data governance with high-quality, representative datasets. As a form of contribution towards meeting these requirements, the \textsc{UniversalCEFR} is an initiative that will allow researchers and developers access to diverse, multilingual, multidimensional CEFR-labeled texts which can be used for designing systems that are representative, explainable, and fair.



\section*{Acknowledgments}
JMI is supported by the National University Philippines and the UKRI Centre for Doctoral Training in Accountable, Responsible, and Transparent AI [EP/S023437/1] of the University of Bath.

HS has received funding from the European Union’s Horizon Europe research and innovation program under Grant Agreement No. 101132431 (iDEM Project). HS also receives support from the Spanish State Research Agency under the Maria de Maeztu Units of Excellence Programme (CEX2021-001195-M) and from the  Departament de Recerca i Universitats de la Generalitat de Catalunya (ajuts SGR-Cat 2021).

DK and FAM have received funding from the Welsh Government as part of the ``Developing a CEFR Predictor for Welsh (2025-26)'' project.

ER is supported by Portuguese national funds through Fundação para a Ciência e a Tecnologia (Reference: UIDB/50021/2020, DOI: 10.54499/UIDB/50021/2020) and by the European Commission (Project: iRead4Skills, Grant number: 1010094837, Topic: HORIZON-CL2-2022-TRANSFORMATIONS-01-07, DOI: 10.3030/101094837).

\bibliography{anthology,custom}

\clearpage
\appendix

\section{Full Data Statistics}
\label{app:full_data_statistics}

Tables~\ref{tab:universalcefr_main_count_level}, \ref{tab:universalcefr_train_count_level}, \ref{tab:universalcefr_dev_count_level}and \ref{tab:universalcefr_test_count_level} report the quantity of CEFR-labeled texts across granularity levels per language, and Tables~\ref{tab:universalcefr_main_count_grade_coverage}, \ref{tab:universalcefr_train_count_grade_coverage}, \ref{tab:universalcefr_dev_count_grade_coverage} and \ref{tab:universalcefr_test_count_grade_coverage} reflect their counterparts in terms of CEFR level coverage. In forming the \textsc{test} split, we randomly sampled CEFR-labeled text instances per language per granularity level, while setting a cap of 200. This allows us to have a sizeable representation of \textsc{UniversalCEFR} while maintaining efficiency for inference with LLMs. In total, we have 4,465 CEFR-labeled instances for \textsc{UniversalCEFR-test}, which is comparable to the general sizes of benchmark test sets from previous works related to language proficiency \cite{naous-etal-2024-readme,zhang-etal-2024-prolex,imperial-tayyar-madabushi-2024-specialex}. For the \textsc{train} and \textsc{dev} sets for fine-tuning and feature-based classification, we split the \textsc{full} subset (minus the \textsc{test} set) into a 90\%-10\% partition, respectively.



\setlength{\tabcolsep}{5pt} 
\begin{table}[t]
\centering
\footnotesize
\begin{tabular}{@{}lrrrr@{}}
\toprule
{\textsc{Lang}} & {\textsc{Sent}} & {\textsc{Para}} & {\textsc{Doc}} & {\textsc{Diag}}\\ \midrule
\textsc{EN}         & 12,826   & 409,362   & 1,837    & 0        \\
\textsc{ES}          & 0        & 713       & 31,355   & 0        \\
\textsc{DE}           & 26,244   & 1,033     & 5,673    & 0        \\
\textsc{NL}            & 0        & 0         & 3,596    & 0        \\
\textsc{CS}            & 0        & 441       & 0        & 0        \\
\textsc{IT}         & 0        & 813       & 0        & 0        \\
\textsc{FR}           & 1,669    & 0         & 344      & 0        \\
\textsc{ET}         & 0        & 420       & 1,277    & 0        \\
\textsc{PT}       & 0        & 1,423     & 0        & 0        \\
\textsc{AR}           & 1,945    & 215       & 0        & 0        \\
\textsc{HI}            & 1,491    & 0         & 0        & 0        \\
\textsc{RU}          & 1,758    & 0         & 0        & 0        \\
\textsc{CY}            & 1,107    & 109       & 41       & 115      \\ \midrule
\textbf{Total}                 & \bf 47,040    & \bf 414,529    & \bf 115      & \bf 44,123    \\ \bottomrule
\end{tabular}
\caption{Data statistics of \textbf{\textsc{UniversalCEFR-full}} in terms of levels (sentence, paragraph, document, dialogue) across the 13 target languages.}
\label{tab:universalcefr_main_count_level}
\end{table}


\setlength{\tabcolsep}{3pt} 
\begin{table}[htbp]
\centering
\footnotesize
\begin{tabular}{@{}lrrrrrr@{}}
\toprule
{\textsc{Lang}}  & {\textsc{A1}}     & {\textsc{A2}}     & {\textsc{B1}}     & {\textsc{B2}}    & {\textsc{C1}}    & {\textsc{C2}}   \\ \midrule
EN       & 173,005 & 119,335 & 59,634 & 20,746               & 7,122 & 675                  \\
ES       & 4577    & 4989    & 4,051  & 3,007                & 1,707 & 0                    \\
DE       & 273     & 13,208  & 12,996 & 346                  & 108   & 308                  \\
NL       & 18      & 93      & 323    & 277                  & 84    & 33                   \\
CS       & 1       & 92      & 77     & 38                   & 2     & 0                    \\
IT       & 17      & 261     & 267    & 1                    & 0     & 0                    \\
FR       & 106     & 302     & 404    & 335                  & 210   & 98                   \\
ET       & 0       & 266     & 406    & 293                  & 215   & 0                    \\
PT       & 204     & 62      & 270    & 59                   & 80    & 0                    \\
AR       & 62      & 207     & 407    & 445                  & 285   & 153                  \\
HI       & 203     & 219     & 223    & 203                  & 182   & 145                  \\
RU       & 327     & 234     & 331    & 256                  & 192   & 69                   \\
CY       & 463     & 332     & 0      & 0                    & 0     & 0                    \\\midrule
\textbf{Total} & \textbf{179,256} & \textbf{139,600} & \textbf{79,389} & \textbf{26,006} & \textbf{10,187} & \textbf{1,481} \\ \bottomrule
\end{tabular}
\caption{Data statistics of \textbf{\textsc{UniversalCEFR-train}} in terms of recognized CEFR levels (A1, A2, B1, B2, C1, C2) across the 13 target languages.}
\label{tab:universalcefr_train_count_grade_coverage}
\end{table}

\setlength{\tabcolsep}{5pt} 
\begin{table}[t]
\centering
\footnotesize
\begin{tabular}{@{}lrrrr@{}}
\toprule
{\textsc{Lang}} & {\textsc{Sent}} & {\textsc{Para}} & {\textsc{Doc}} & {\textsc{Diag}}\\ \midrule
\textsc{EN}         & 12,826   & 409,362   & 1,837    & 0        \\
\textsc{ES}          & 0        & 713       & 31,355   & 0        \\
\textsc{DE}           & 26,244   & 1,033     & 5,673    & 0        \\
\textsc{NL}            & 0        & 0         & 3,596    & 0        \\
\textsc{CS}            & 0        & 441       & 0        & 0        \\
\textsc{IT}         & 0        & 813       & 0        & 0        \\
\textsc{FR}           & 1,669    & 0         & 344      & 0        \\
\textsc{ET}         & 0        & 420       & 1,277    & 0        \\
\textsc{PT}       & 0        & 1,423     & 0        & 0        \\
\textsc{AR}           & 1,945    & 215       & 0        & 0        \\
\textsc{HI}            & 1,491    & 0         & 0        & 0        \\
\textsc{RU}          & 1,758    & 0         & 0        & 0        \\
\textsc{CY}            & 1,107    & 109       & 41       & 115      \\ \midrule
\textbf{Total}                 & \bf 47,040    & \bf 414,529    & \bf 115      & \bf 44,123    \\ \bottomrule
\end{tabular}
\caption{Data statistics of \textbf{\textsc{UniversalCEFR-train}} in terms of levels (sentence, paragraph, document, dialogue) across the 13 target languages.}
\label{tab:universalcefr_train_count_level}
\end{table}


\setlength{\tabcolsep}{3pt} 
\begin{table}[htbp]
\centering
\footnotesize
\begin{tabular}{@{}lrrrrrr@{}}
\toprule
{\textsc{Lang}}  & {\textsc{A1}}     & {\textsc{A2}}     & {\textsc{B1}}     & {\textsc{B2}}    & {\textsc{C1}}    & {\textsc{C2}}   \\ \midrule
EN       & 19,449 & 13,151 & 6,643 & 2,384                & 797 & 85 \\
ES       & 1535   & 1226   & 904   & 471                  & 285 & 0  \\
DE       & 32     & 2,494  & 2,392 & 60                   & 13  & 41 \\
NL       & 6      & 70     & 235   & 230                  & 99  & 32 \\
CS       & 0      & 14     & 9     & 6                    & 0   & 0  \\
IT       & 3      & 33     & 23    & 1                    & 0   & 0  \\
FR       & 13     & 30     & 39    & 43                   & 20  & 12 \\
ET       & 0      & 19     & 52    & 21                   & 25  & 0  \\
PT       & 61     & 213    & 50    & 144                  & 19  & 61 \\
AR       & 7      & 26     & 56    & 53                   & 35  & 15 \\
HI       & 22     & 30     & 20    & 16                   & 12  & 13 \\
RU       & 34     & 23     & 25    & 34                   & 21  & 9  \\
CY       & 67     & 44     & 0     & 0                    & 0   & 0  \\  \midrule
\textbf{Total} & \textbf{21,229} & \textbf{17,373} & \textbf{10,448} & \textbf{3,463} & \textbf{1,326} & \textbf{268} \\ \bottomrule
\end{tabular}
\caption{Data statistics of \textbf{\textsc{UniversalCEFR-dev}} in terms of recognized CEFR levels (A1, A2, B1, B2, C1, C2) across the 13 target languages.}
\label{tab:universalcefr_dev_count_grade_coverage}
\end{table}

\setlength{\tabcolsep}{5pt} 
\begin{table}[t]
\centering
\footnotesize
\begin{tabular}{@{}lrrrr@{}}
\toprule
{\textsc{Lang}} & {\textsc{Sent}} & {\textsc{Para}} & {\textsc{Doc}} & {\textsc{Diag}}\\ \midrule
EN & 1,274 & 40,980 & 0 & 255   \\
ES & 0     & 51     & 0 & 4,370 \\
DE & 4,168 & 79     & 0 & 785   \\
NL & 0     & 0      & 0 & 672   \\
CS & 0     & 29     & 0 & 0     \\
IT & 0     & 60     & 0 & 0     \\
FR & 146   & 0      & 0 & 11    \\
ET & 0     & 19     & 0 & 98    \\
PT & 0     & 548    & 0 & 0     \\
AR & 188   & 4      & 0 & 0     \\
HI & 113   & 0      & 0 & 0     \\
RU & 146   & 0      & 0 & 0     \\
CY & 111   & 0      & 0 & 0     \\ \midrule
\textbf{Total} & \textbf{6,146}       & \textbf{41,770}      & \textbf{0}           & \textbf{6,191}   \\ \bottomrule
\end{tabular}
\caption{Data statistics of \textbf{\textsc{UniversalCEFR-dev}} in terms of levels (sentence, paragraph, document, dialogue) across the 13 target languages.}
\label{tab:universalcefr_dev_count_level}
\end{table}


\setlength{\tabcolsep}{8pt} 
\begin{table}[t]
\centering
\footnotesize
\begin{tabular}{@{}lrrrrrr@{}}
\toprule
{\textsc{lang}}  & {\textsc{A1}}     & {\textsc{A2}}     & {\textsc{B1}}     & {\textsc{B2}}    & {\textsc{C1}}    & {\textsc{C2}}   \\ \midrule
EN & 107 & 114 & 132 & 129 & 83 & 35 \\
ES & 49  & 58  & 140 & 108 & 45 & 0  \\
DE & 14  & 264 & 238 & 67  & 9  & 8  \\
NL & 4   & 21  & 69  & 77  & 22 & 7  \\
CS & 0   & 82  & 79  & 37  & 2  & 0  \\
IT & 9   & 87  & 104 & 0   & 0  & 0  \\
FR & 32  & 57  & 132 & 100 & 63 & 16 \\
ET & 0   & 110 & 130 & 93  & 67 & 0  \\
PT & 49  & 50  & 47  & 30  & 13 & 11 \\
AR & 12  & 26  & 162 & 145 & 40 & 15 \\
HI & 38  & 34  & 42  & 42  & 28 & 16 \\
RU & 41  & 36  & 52  & 35  & 24 & 12 \\
CY & 233 & 232 & 0   & 0   & 0  & 0  \\
\midrule
\textbf{Total} & \textbf{588} & \textbf{1,171} & \textbf{1,327} & \textbf{863} & \textbf{396} & \textbf{120} \\ \bottomrule
\end{tabular}
\caption{Data statistics of \textbf{\textsc{UniversalCEFR-test}} in terms of recognized CEFR levels (A1, A2, B1, B2, C1, C2) across the 13 target languages.}
\label{tab:universalcefr_test_count_grade_coverage}
\end{table}

\begin{table}[t]
\centering
\footnotesize
\begin{tabular}{@{}lrrrr@{}}
\toprule
{\textsc{Lang}} & {\textsc{Sent}} & {\textsc{Para}} & {\textsc{Doc}} & {\textsc{Diag}}\\ \midrule
\textsc{EN}         & 200   & 200    & 200     & 0        \\
\textsc{ES}          & 0        & 200        & 200    & 0        \\
\textsc{DE}           & 200    & 200      & 200     & 0        \\
\textsc{NL}            & 0        & 0         & 200     & 0        \\
\textsc{CS}            & 0        & 200        & 0        & 0        \\
\textsc{IT}         & 0        & 200        & 0        & 0        \\
\textsc{FR}           & 200     & 0         & 200       & 0        \\
\textsc{ET}         & 0        & 200        & 200     & 0        \\
\textsc{PT}       & 0        & 200      & 0        & 0        \\
\textsc{AR}           & 200     & 200        & 0        & 0        \\
\textsc{HI}            & 200     & 0         & 0        & 0        \\
\textsc{RU}          & 200     & 0         & 0        & 0        \\
\textsc{CY}            & 200     & 109       & 41       & 115      \\ \midrule
\textbf{Total}                 & \bf 1,400    & \bf 1,709    & \bf 1,241     & \bf 115    \\ \bottomrule
\end{tabular}
\caption{Data statistics of \textbf{\textsc{UniversalCEFR-test}} in terms of levels (sentence, paragraph, document, dialogue) across the 13 target languages.}
\label{tab:universalcefr_test_count_level}
\end{table}

\section{Coverage of Large Language Models}
\label{app:language_coverage_llms}
In Table~\ref{tab:language_coverage_llms}, we map each model's language coverage or language support based on its respective release papers and publications. Language support means what specific languages have been added and in substantial quantities in a model's training data (e.g., multilingual Wikipedia data dumps for pretraining XLM-R \cite{conneau-etal-2020-unsupervised}).

\setlength{\tabcolsep}{6pt} 
\begin{table*}[!t]
\centering
\small
\begin{tabular}{@{}lcccccccccccccr@{}}
\toprule
Model & \textsc{EN} & \textsc{ES} & \textsc{DE} & \textsc{NL} & \textsc{CS} & \textsc{IT} & \textsc{FR} & \textsc{ET} & \textsc{PT} & \textsc{AR} & \textsc{HI} & \textsc{RU} & \textsc{CY} & Tally \\ 
\midrule
\textcolor{teal}{\textsc{Gemma1}}    & \checkmark &            &            &            &            &            &            &            &            &            &            &            &           &    1/1   \\
\textcolor{violet}{\textsc{Gemma3}}    & \checkmark & \checkmark & \checkmark & \checkmark & \checkmark & \checkmark & \checkmark & \checkmark & \checkmark & \checkmark & \checkmark & \checkmark & \checkmark &    13/140   \\
\textcolor{violet}{\textsc{EuroLLM}}   & \checkmark & \checkmark & \checkmark & \checkmark & \checkmark & \checkmark & \checkmark & \checkmark & \checkmark & \checkmark & \checkmark & \checkmark &            &    12/35   \\
\textcolor{teal}{\textsc{ModernBERT}}& \checkmark &            &            &            &            &            &            &            &            &            &            &            &            &    1/1   \\
\textcolor{violet}{\textsc{EuroBERT}}  & \checkmark & \checkmark & \checkmark & \checkmark &            & \checkmark & \checkmark &            & \checkmark & \checkmark & \checkmark & \checkmark &            &   10/15    \\
\textcolor{violet}{\textsc{XLM-R}}     & \checkmark & \checkmark & \checkmark & \checkmark & \checkmark & \checkmark & \checkmark & \checkmark & \checkmark & \checkmark & \checkmark & \checkmark & \checkmark &   13/100    \\
\bottomrule
\end{tabular}
\caption{Mapping of language coverage of training data used for the six large, pretrained language models in the model evaluation paradigm in Section~\ref{sec:cefr_level_classification}. Models in \textcolor{teal}{\bf teal} are English-centric (trained primarily with English data), and models in \textcolor{violet}{\bf purple} are multilingual (trained with massive multilingual data). We referred to each model's corresponding release papers and publications for information on their supported languages. Note that the documentation of \textsc{Gemma3} indicates it has been trained with 140+ languages. Thus, we loosely consider it to cover all 13 languages in \textsc{UniversalCEFR}. The tally column indicates \{\texttt{lang\_covered}/\texttt{lang\_seen}\}. For example, \textsc{EuroBERT} covers 10 of the languages in the current \textsc{UniversalCEFR} the 15 languages it supports.}
\label{tab:language_coverage_llms}
\end{table*}

\section{Language-Specific Analysis}
\label{app:language_specific_analysis}
We provide in-depth analysis of model performances from the experiments in Section~\ref{sec:cefr_level_classification} across multiple dimensions of \textsc{UniversalCEFR} on results for selected languages that we are qualified to interpret. \\

\noindent \textbf{English}. 
Analysis of model performance shows that using fine-tuned models and linguistic feature-based classification (62\%-75\%) obtains the best performance compared to prompting with instruction-tuned LLMs (19\%-28\%). However, these models tend to provide distinct patterns of specific CEFR labels. For the prompting setup,  Gemma1, Gemma3, and EuroLLM models tend to give labels within the A1 and B1 range, while fine-tuned and feature-based models tend to lean towards the B1 and B2 range. For the pre-trained and instruction-tuned models, this finding may be tied to A1 and B2 being the most common CEFR level band of most general-purpose texts found online, where the sources of the data from which these models are trained. For feature-based models, we note the potential effect of training and test data having higher instance counts for these level bands than A1, C1, and C2. Regarding model scale, upgraded versions from similar model families perform better than their previous versions, echoing previous findings in literature \cite{imperial-tayyar-madabushi-2024-specialex}. This is particularly evident in Gemma3 being 12B in size and trained with massively multilingual data in 140+ languages and obtaining 28\% in weighted F1 compared to Gemma1, which is 7B in size and English-centric, obtaining 21.8\%. We note a potential \textit{default effect} in using these models where additional specific CEFR descriptor information is not needed if the texts being evaluated are in English, due to the majority of data in the context of CEFR that is reflected in the training data being English.\\

\noindent \textbf{Spanish}. 
Fine-tuned models outperform other setups, with feature-based approaches, especially Random Forest, achieving reasonable comparative performance. Moreover, multilingual models provide noticeable performance gains when compared to the English-only model. As per prompting strategy, for smaller multilingual models the language-specific prompt seems to play a role in improving the performance as it also does for the Gemma1 English-only model, however, the Gemma3 with 12B parameter is not affected by this, and it has been able to produce the best results of the LLMs (plus more sophisticated prompting strategies). As for the granularity of the input, models perform noticeably better at the document level than at the paragraph level, indicating that longer contexts are easier to classify than short ones.  Finally, it is worth reporting a noticeable error of  Gemma1: the prediction of C2 grade level,  which does not exist in the Spanish dataset.  \\

\noindent \textbf{Hindi}. Both the Gemma models perform poorly compared to the fine-tuned XLM-R and the Random Forest variants and tend to classify most Hindi test items as A1 or A2. For example, Gemma1 puts 57\% of Hindi test samples as A1, whereas there are only 19\% of the test samples labeled as A1 in the gold standard labels. This is in line with the general trend noticed in Section~\ref{subsec:granularity}, as the Hindi subset is entirely sentence-level. The distribution is closer to the Gold distribution for the fine-tuned and feature-engineered models. XLM-R fine-tuned models give the best performance amongst all models for Hindi, both in terms of exact category prediction and in terms of the degree of error (i.e., being within 1 level above or below the correct level). Finally, we looked at the correlation between a simple approximation of text length (calculated as the number of space-separated tokens), a commonly used variable in such automated language assessment approaches in NLP research, and the CEFR gold labels, as well as model-predicted labels, after converting them to a numeric scale. There was a high correlation between text length and the gold labels (0.7), which was also seen with the XLM-R model (0.74) and the Random Forest models (0.77). However, the Gemma models only had correlations of 0.44 and 0.54, respectively, with text length. However, considering that the Hindi subset only has sentence-level annotations without a larger context, it may be challenging to achieve further consistency with the gold standard labels, given the size of the annotated dataset. Future research should expand the available CEFR-graded resources both in terms of quantity as well as granularity for the language. \\

\noindent \textbf{Russian}. The Russian results follow the broad patterns reported in the paper, but their rich inflectional morphology and their comparatively limited training data amplify several effects. Gemma1 (34.8\%) greatly over-predicts texts as beginner-level (only 5\% of texts had predictions above B1), confirming the overall trend that small, English-centric LLMs struggle most with morphologically rich languages. Gemma3 (37.4\%) partially corrects this, but still massively under-predicts B2 and C2. XLM-R (49.6\%) mirrors the gold distribution most faithfully, possibly because its multilingual vocabulary gives it better coverage of Russian inflectional morphology, a pattern also seen for other highly inflected languages such as Czech. The two Random Forest models (47.2\% and 47.8\%) under-predict A2 and C2 but otherwise match the gold shape, showing that handcrafted lexical and morpho-syntactic features capture useful Russian-specific signals even with limited data. Subword-level multilingual models (XLM-R) or explicit morpho-syntactic features (RF) are best suited to capture the meanings and relations between Russian words. Text length appears to be a false friend; although it does correlate highly with readability (r=0.65), it also appears to be the source of many errors; top-performing model outputs had text length correlations as high as 0.73. Since this experiment with Russian is limited to sentence-level readability, comparison with previous research on Russian readability assessment is not straightforward. However, the weighted F1 (49.6\%) of the best-performing model (XLM-R) is below state-of-the-art results for longer texts, including 67\%  \citep{reynolds-2016-insights}, 74\% \citep{solnyshkina2018readability}, and 78\% \citep{blinova2022hybrid}. Most likely, this difference is partly due to the absence of Russian-specific morphosyntactic features that have been highly informative in previous studies' models. \\  

\noindent \textbf{Portuguese}.
Comparing the different setups, we can see that the results for Portuguese follow the global tendency, with fine-tuned models achieving the highest performance, followed by feature-based models, and with prompting taking the last place. Although this study only covers paragraph-level learner data for Portuguese, similar patterns were observed on reference data~\cite{ribeiro2024avaliacao}. However, comparing the results with those of other languages and, particularly, those with paragraph-level learner data, we can see that Portuguese is the language with the lowest performance ($\approx$33.5\%). Several factors may contribute to this outcome. For instance, Portuguese is one of the languages with the least available training data, and the distribution of proficiency labels is right-skewed (especially in COPLE2). Furthermore, the data consists of texts written by learners from a wide range of L1 backgrounds with generally low proficiency. This makes it more difficult for models to identify consistent patterns due to strong L1 interference and low coverage. Overall, both fine-tuned and feature-based models seem to be unable to distinguish between sublevels, with most examples of both A levels being predicted as A1, and the remainder (mostly examples of the B levels) as B1. On the positive side, contrary to what was observed for other languages, the models do not seem to be influenced by text length, with the predictions of XML-R having a correlation of just 0.39 with that feature. The prompting approaches lead to a bias towards the prediction of levels A2 and B1, with the top performer among these approaches (Gemma3 with \textsc{En-Write} prompt) predicting A2 for 28\% of the examples and B1 for 62\%. Notably, when using the more descriptive prompts, the Gemma 1 model outperformed EuroLLM, in spite of having fewer parameters and not being specifically trained on Portuguese data. \\

\noindent \textbf{French}. The French corpus and our analysis are divided into sentence-level and document-level data. The sentence-level set contains 1,668 sentences ranging from A1 to C2, while the document-level set includes 344 documents from A1 to C1, with an intense concentration at the B levels (75\% of the data falls within B1 and B2). In line with the other languages, XLM-R is the most consistent model and achieves the best global performance in every setting. Random Forest (RF) with all features fluctuates more in overall performance, dropping notably in the document-level task, but retains some consistency in terms of which proficiency levels it performs best or worst on. RF with top features performs inconsistently overall but achieves the best results on the document-level task. However, it shows instability in class-level performance, with changes in which levels are most accurately predicted. Among the prompt-based models, Gemma3 is more stable than Gemma1, but both remain below the performance of XLM-R and RF, showing a weaker performance in the LLMs (Gemma1 and Gemma3). Gemma1, in particular, is the least consistent model, with highly variable class-level performance and occasional zero F1 scores for some levels in specific setups. The Gemma1 results are likely due to the lack of French documents during the training of this model. Across all models, prediction is generally more reliable for intermediate levels (A2–B2), while C-level predictions remain the most challenging. Fine-tuning has the clear advantage: the fine-tuned XLM-R achieves the highest accuracy across all evaluation set-ups, making it the most reliable in correctly predicting gold labels. It consistently outperforms all other models, both at the sentence and document levels. This is consistent with previous experiments on French \cite{yancey2021investigating, ngo-parmentier-2023-towards, wilkens2024exploring}, although our performance is slightly lower than in those studies. Prompting is the least effective: both Gemma1 and Gemma3, used in a prompt-based setting, show the lowest prediction accuracy, often failing to identify the correct labels, especially at the extremes of the proficiency scale (A1, C1, and C2 levels). Traditional supervised classifiers (Random Forest) perform moderately well, consistently outperforming the prompt-based models but still lagging behind the fine-tuned model. The feature-based models had a particularly poor performance on C1 and C2 levels. This is likely due to a lack of specialized features for those proficiency levels. Moreover, their performance varies by set-up, with some gains at the document level but noticeable drops elsewhere. Nevertheless, the two RF flavours had similar results. In summary, fine-tuning yields the best predictions, followed by traditional supervised learning, while prompting underperforms in this task.\\

\noindent \textbf{German}. 
For German, the fine-tuned models (>70\%) have been shown to outperform all other approaches, such as feature-based ($\approx$50\%-65\%) and prompting ($\approx$38\%-46\%), despite the presence of unbalanced CEFR levels in both the training and test data. The findings derived from the English-only and multilingual models, including fine-tuning and prompting methodologies, exhibit no notable difference. This may be due to the similarities between English and German, both of which are West Germanic languages. Alternatively, the great transferability of the fine-tuned English-only model may also be due to the large amount of German training data available (27,000 training samples). The feature-based models performed second best and were still able to compete with the fine-tuned models to some extent. This is surprising, given that a previous analysis showed that the features only exhibited low correlations with CEFR levels (see Section~\ref{app:appendix-feature-correlation-analysis}). Proficiency assessment for German appears to require certain idiosyncratic features. For example, the feature covering the maximum distance between words in a dependency tree showed a high feature importance only for German, reflecting the language's free word order and long-distance dependencies. 
For the prompting setup, the multilingual Gemma3 model performed, achieving good results for lower CEFR levels, but underpredicting higher levels. By contrast, Gemma1 significantly overpredicts level A1 (250 against 14 from the gold labels), resulting in poorer performance on average and across the other levels. One deceptive indicator might be the length of the texts to be classified, as reflected by the strong correlation between text length and Gemma1's predictions ($r$=0.61). When comparing the prompting setups with regard to language-specific task descriptions, no clear trend emerges across all three LLMs, mirroring the difficulty of prompt engineering for a complex task such as multi-lingual proficiency classification. \\

\noindent \textbf{Arabic}. 
Across the 400 Arabic test items, Gemma1 tends to over-predict lower CEFR levels, assigning 31 items to A1 while only 12 are from the true labels, and 90 to A2 against 26. There is also a tendency to under-predict C1, with 18 predictions against 40 from the true labels, resulting in the highest average grade deviation of 1.0. In contrast, XLM-R and both Random Forest variants distributed their predictions more evenly overall, with XLM-R achieving the smallest average grade deviation of 0.75. In terms of granularity, the Arabic subset is split into sentence-level, reference data, and paragraph-level learner data. For the sentence-level reference texts, XLM-R ($\approx$55\%) and Random Forest models from the two linguistic feature setups ($\approx$49.3\%-51.2\%) outperform both Gemma1 and Gemma3 models through prompting ($\approx$16.5\%-32\%). However, with paragraph-level learner texts, Gemma3 leads the evaluation ($\approx$41\%). At the same time, XLM-R and the Random Forest models fall behind ($\approx$32\%), possibly due to the Arabic data used in the training split, which are entirely sentence-level. In contrast, the Gemma3 model has most likely seen diverse online Arabic data.



\section{Standardized Dataset Fields}
We present the standardized JSON format used as a template when processing all qualified datasets in \textsc{UniversalCEFR}. This structured format ensures flexibility and interoperability into other formats accepted and used by the AI community, including Huggingface and Croissant. Moreover, this format captures the dimensions that are essential to each instance of CEFR-labeled text, including format or granularity, category, license, and language.


\begin{table*}[htbp]
\centering
\begin{tabular}{p{0.20\linewidth}  p{0.7\linewidth}}
\toprule
\textbf{Field} & \textbf{Description} \\
\midrule

\texttt{title} & The unique title of the text retrieved from its original corpus (\texttt{NA} if there are no titles such as CEFR-assessed sentences or paragraphs). \\
\midrule

\texttt{lang} & The source language of the text in ISO 638-1 format (e.g., \texttt{en} for English). \\
\midrule

\texttt{source\_name} & The source dataset name where the text is collected as indicated from their source dataset, paper, and/or documentation (e.g., \texttt{cambridge-exams} from \citet{xia-etal-2016-text}). \\
\midrule

\texttt{format} & The format of the text in terms of level of granularity as indicated from their source dataset, paper, and/or documentation. The recognized formats are the following: [\texttt{document-level}, \texttt{paragraph-level}, \texttt{discourse-level}, \texttt{sentence-level}]. \\
\midrule

\texttt{category} & The classification of the text in terms of who created the material. The recognized categories are \texttt{reference} for texts created by experts, teachers, and language learning professionals and \texttt{learner} for texts written by language learners and students.  \\
\midrule

\texttt{cefr\_level} & The CEFR level associated with the text. The six recognized CEFR levels are the following: [\texttt{A1, A2, B1, B2, C1, C2}]. A small fraction (<1\%) of text in \textsc{UniversalCEFR} contains unlabelled text, texts with plus signs (e.g., \texttt{A1+}), and texts with no level indicator (e.g., \texttt{A, B}).    \\
\midrule

\texttt{license} & The licensing information associated with the text (\texttt{Unknown} if not stated). \\
\midrule

\texttt{text} & The actual content of the text itself. \\

\bottomrule
\end{tabular}
\caption{The structured JSON fields with descriptions and examples used as the standardized uniform format for building the \textsc{UniversalCEFR} dataset. All instances validated from the collection of CEFR-labelled corpora conform to this format.}
\label{tab:json_fields}
\end{table*}

\section{Full Linguistic Feature Analysis}
\label{app:appendix-features}

\begin{table}[t]
\small
\begin{tabularx}{\columnwidth}{lX}
\toprule
\textsc{Category} & \textsc{Feature Name} \\
\toprule
\multirow{9}{*}{Length} & doc\_num\_sents \\
 & doc\_num\_tokens \\
 & num\_characters \\
 & num\_characters\_per\_sentence \\
 & num\_characters\_per\_word \\
 & num\_syllables\_in\_sentence \\
 & num\_syllables\_per\_sentence \\
 & num\_syllables\_per\_word \\
 & num\_words \\
\multirow{2}{*}{Lexical} & average\_pos\_in\_freq\_table \\
 & lexical\_complexity\_score \\
\multirow{4}{*}{Morphosyntactic} & ratio\_Tense\_Past \\
 & ratio\_of\_determiners \\
 & ratio\_of\_numerals \\
 & ratio\_of\_pronouns \\
\multirow{2}{*}{Psycholinguistic} & concreteness \\
 & imagebility \\
\multirow{2}{*}{Readability} & sentence\_fkgl \\
 & sentence\_fre \\
\multirow{4}{*}{Syntactic} & avg\_distance\_between\_words \\
 & average\_length\_VP \\
 & parse\_tree\_height \\
 & ratio\_of\_coordinating\_clauses\\\bottomrule
\end{tabularx}
\caption{List of linguistic features occurring in the top 10 of at least three languages. We use this list for the \textsc{TopFeatures} subset used in the experiment result in Table~\ref{tab:main_results_grouped}.}
\label{tab:topfeature}
\end{table}

\subsection{All Linguistic Features}
\label{app:appendix-allfeatures}
Overall, we have extracted \textbf{100 diverse linguistic features} which can be grouped into morphosyntactic (62), syntactic (18), length-based (11), lexical (4), readability (2), psycholinguistic (2), and discourse (1). The full list of features, including short descriptions, is available in Appendix \ref{app:appendix-features}.
We extracted a diverse set of 100 linguistic features based on sentence-based linguistic annotation with \texttt{spacy}~\cite{ines_montani_2023_10009823} and \texttt{stanza}~\cite{qi-etal-2020-stanza}, including tokenization, part-of-speech tagging, and dependency parsing performed. Additionally, we use \texttt{fasttext} embeddings~\cite{grave-etal-2018-learning}, \texttt{pyphen} for hyphenation~\cite{pyphen} and \texttt{MEGA.HR} crossling lexicon\footnote{\url{https://www.clarin.si/repository/xmlui/handle/11356/1187}} for imageability and concreteness~\cite{11356/1187}. Most of the features have already been implemented in the \texttt{text-simplification-evaluation (TSEval)} package\footnote{\url{https://github.com/facebookresearch/text-simplification-evaluation}} (see \citet{martin-etal-2018-reference} for the original version and \citet{stodden-kallmeyer-2020-multi} for the multilingual version).

In \autoref{tab:allfeature-2}, we provide an overview of all features including a short description, resources used, and correlation with the CEFR level.

\subsection{Top Linguistic Features}
\label{app:appendix-topfeatures}
To extract the top linguistic features (\textsc{TopFeats}), we selected those that are present in the top 10 ranked most important features for at least three languages. Using this criteria, we came up with a list of 23 linguistic features as reported in Table~\ref{tab:topfeature} which was then used in the experiment result in Table~\ref{tab:main_results_grouped}.

\subsection{Linguistic Correlation Analysis} 
\label{app:appendix-feature-correlation-analysis}
In the following, we describe some insights into linguistic diversity of the UniversalCEFR data by correlation analysis between the features and the CEFR levels.

\paragraph{Correlation Across All Languages.} Considering the absolute Spearman correlation between the features and the CEFR level (selecting values with $p < 0.05$ and $\rho > 0.3$ on average across all languages), the strongest associations were found in length-based measures, such as characters per sentence and syllables per sentence. Several grammatical complexity features, including parse tree height and phrase length, showed moderate correlations. Readability indices (FKGL and Flesch Reading Ease) also displayed moderate correlations in the expected direction. Psycholinguistic features, such as concreteness and imageability, were negatively correlated with proficiency, indicating a shift toward more abstract language at higher levels. Finally, morphosyntactic features regarding voice, tense, and number showed moderate but consistent correlations, supporting their relevance in reflecting syntactic development.

\paragraph{Correlation By CEFR Level.} To assess the consistency of feature relevance across languages, we examined the number of features with significant correlations ($p < 0.05$) with CEFR levels per language. The results revealed notable variations. Languages such as Czech (\textsc{cs}), Estonian (\textsc{et}), and Italian (\textsc{it}) showed a high number of relevant features, suggesting strong alignment between the selected linguistic features and CEFR progression in these languages. English (\textsc{en}), Spanish (\textsc{es}), French (\textsc{fr}), Hindi (\textsc{hi}), and Russian (\textsc{ru}) showed moderate coverage, with a reasonable number of features exceeding the 0.3 correlation threshold. In contrast, Arabic (\textsc{ar}), Dutch (\textsc{nl}), and Portuguese (\textsc{pt}) exhibited weak coverage, while Welsh (\textsc{cy}) and German (\textsc{de}) had very few or no features with relevant correlations, indicating a limited match between the current feature set and CEFR levels for those languages. Furthermore, a few features are only relevant for a few languages, e.g., the translative case for only Estonian, negative verb polarity for only Czech, or genitive case for only Czech, Estonian, and Russian. This variability highlights the influence of language-specific properties on the effectiveness of general feature-based models for proficiency prediction.

\paragraph{Point-Biserial Correlation.} A point-biserial correlation analysis by CEFR level revealed that most features exhibit only weak correlations, suggesting limited discriminative power when isolating individual CEFR bands. Interestingly, the absolute correlation values tend to be strongest at the A1 level, particularly for psycholinguistic features such as imageability ($\rho = 0.48$) and concreteness ($\rho = 0.46$), as well as punctuation-related measures. This suggests that certain surface-level and lexical-semantic features may be especially informative at the lowest proficiency level. A notable case is the feature of word length in characters, which shows a negative correlation at A1 ($\rho = -0.45$), becomes neutral at A2, and shifts to a positive correlation at B1 and higher levels. This pattern may reflect increasing lexical complexity with proficiency. Similarly, features related to syntactic structure, such as the ratio of past tense verbs and phrase length, generally shift from weak negative to weak positive correlations as proficiency increases, indicating progressive syntactic development. Overall, the directionality of several features suggests dynamic usage patterns across CEFR bands, even if the correlation strengths remain modest.

\section{Hyperparameter Values}
We detail the hyperparameter values used for fine-tuning pretrained (\textsc{ModernBERT}, \textsc{EuroBERT}, and \textsc{XLM-R}) and instruction-tuned language models (\textsc{Gemma1}, \textsc{Gemma3}, and \textsc{EuroLLM}) in Tables~\ref{tab:fine-tuning_hyperparams} and \ref{tab:promtping_hyperparams}, respectively.

\begin{table}[!t]
  \centering
  \small
  \begin{tabular}{ll}
    \toprule
    \textsc{Hyperparameter}                & \textsc{Value}                 \\
    \midrule
    Learning rate                          & $3.6 \times 10^{-5}$           \\
    Train batch size                       & 2                              \\
    Evaluation batch size                  & 3                              \\
    Random seed                            & 42                             \\
    Gradient accumulation steps            & 16                             \\
    Total effective batch size             & 32                             \\
    Optimizer                              & \texttt{adamw\_torch\_fused} \\
    \quad Betas                             & (0.9,\,0.999)                 \\
    \quad Epsilon                           & $10^{-8}$                     \\
    Learning‐rate scheduler                & linear                         \\
    Warm‐up ratio                          & 0.1                            \\
    \bottomrule
  \end{tabular}
  \caption{Hyperparameter values used for fine-tuning pretrained language models.}
  \label{tab:fine-tuning_hyperparams}
\end{table}

\begin{table}[!t]
  \centering
  \small
  \begin{tabular}{ll}
    \toprule
    \textsc{Hyperparameter}                & \textsc{Value}                 \\
    \midrule
    Sampling                          & False           \\
    Max New Tokens                       & 10                              \\
    Data Type                  & \texttt{torch.bfloat16}                            \\
    GPU      & 4 x NVIDIA RTX A5000 (24GB)   
    \\                 \bottomrule
  \end{tabular}
  \caption{Hyperparameter values and GPU information used for prompting instruction-tuned models.}
  \label{tab:promtping_hyperparams}
\end{table}



\begin{table}[!t]
\centering
\small
\begin{tabular}{lrrrr} \toprule
\textsc{Lang} & \textsc{Sent} & \textsc{Para} & \textsc{Doc}& \textsc{Overall}\\ \midrule
\textsc{ar}  & \textbf{55.7} & 32.6 & -& 43.1 \\
\textsc{cy}  & \textbf{86.9} & 72.5  & 61.5 & 72.7 \\
\textsc{cs}  & -& 68.8 & -& 68.8 \\
\textsc{de}  & 65.4 & 71.1 & \textbf{83.4}  & 73.2 \\
\textsc{en}  & 68.3 & \textbf{100.0}  & 57.6 & 75.5 \\
\textsc{es}  & -& 40.6 & \textbf{98.0} & 69.69 \\
\textsc{et}  & -& \textbf{93.6} & 84.0  & 88.9 \\
\textsc{fr}  & \textbf{57.6} & - & 44.2  & 51.7 \\
\textsc{hi}  & 52.9 &  -  & -& 52.9 \\
\textsc{it}  & -& 83.3 & -& 83.3 \\
\textsc{nl}  &  &   - & 59.0  & 59.0 \\
\textsc{pt}  & -& 29.2 & -& 29.2 \\
\textsc{ru}  & 49.6  & -& -& 49.6 \\ \bottomrule
\end{tabular}
\caption{Weighted F1 scores for the fine-tuned XLM-R (top model across all setups) performance on the \textsc{UniversalCEFR-test}, classified by the granularity levels of the data.}
\label{tab:granular_lang_perf}
\end{table}

\newpage

\setlength{\tabcolsep}{5pt} 
\begin{sidewaystable*}[!htbp]
\centering
\scriptsize
\begin{tabular}{llllll}
\hline
\textbf{Category} & \textbf{Feature} & \textbf{Short Description} & \textbf{Resource} & \begin{tabular}[c]{@{}l@{}}\textbf{Corr.} $\rho$ \\ \textbf{(avg.)}\end{tabular} & \begin{tabular}[c]{@{}l@{}}\textbf{Corr. }$\rho$ \\ \textbf{(SD)} \end{tabular} \\ \hline
\rowcolor{tab-lightgrey}
Discourse & ratio\_referential & Ratio of referential tokens to all tokens based on dependency tree relations & SpaCy, Stanza, TSEval & 0.1278 & 0.09 \\
\multirow{11}{*}{Length} & doc\_num\_sents & Number of sentences per document / text & SpaCy, Stanza, & 0.1819 & 0.18 \\
 & doc\_num\_tokens & Number of tokens per document / text & SpaCy, Stanza, & 0.5041 & 0.22 \\
 & num\_characters & Number of characters per document / text & SpaCy, Stanza, TSEval & 0.5224 & 0.19 \\
 & num\_characters\_per\_sentence & Number of characters per sentence & SpaCy, Stanza, TSEval & 0.5301 & 0.17 \\
 & num\_characters\_per\_word & Number of characters per word & SpaCy, Stanza, TSEval & 0.3895 & 0.16 \\
 & num\_sentences & Number of sentences per document / text & SpaCy, Stanza, TSEval & -0.0324 & 0.11 \\
 & num\_syllables\_in\_sentence & Number of syllables in document / text & SpaCy, Stanza, pyphen, TSEval & 0.525 & 0.19 \\
 & num\_syllables\_per\_sentence & Number of syllables per sentence & SpaCy, Stanza, pyphen, TSEval & 0.4634 & 0.23 \\
 & num\_syllables\_per\_word & Number of syllables per word & SpaCy, Stanza, pyphen, TSEval & 0.3924 & 0.16 \\
 & num\_words & Number of tokens per document / text & SpaCy, Stanza, TSEval & 0.479 & 0.18 \\
 & num\_words\_per\_sentence & Number of tokens per sentence & SpaCy, Stanza, TSEval & 0.4863 & 0.16 \\
\rowcolor{tab-lightgrey}
& average\_pos\_in\_freq\_table & Average frequency rank of tokens in FastText embeddings & SpaCy, Stanza, TSEval & -0.0907 & 0.22 \\
\rowcolor{tab-lightgrey}
 & lexical\_complexity\_score & Lexical complexity based on ranks of FastText embeddings & SpaCy, Stanza, FastText, TSEval & 0.0649 & 0.13 \\
\rowcolor{tab-lightgrey}
 & type\_token\_ratio & Type-token-Ratio & SpaCy, Stanza, TSEval & -0.3364 & 0.16 \\
\rowcolor{tab-lightgrey}
\multirow{-4}{*}{Lexical} & max\_pos\_in\_freq\_table & Maximum frequency rank of tokens in FastText embeddings & SpaCy, Stanza, TSEval & 0.2014 & 0.07 \\
\multirow{2}{*}{Psycholinguistic} & concreteness & Concreteness of words based on MEGAHR and FastText-Embeddings & MEGA.HR crossling & -0.4254 & 0.24 \\
 & imagebility & imagebility of words based on MEGAHR and FastText-Embeddings & MEGA.HR crossling & -0.3962 & 0.24 \\
\rowcolor{tab-lightgrey}
& sentence\_fkgl & Flesch-Kincaid-Grading-Level, designed for English & SpaCy, Stanza, TSEval & 0.4738 & 0.19 \\
\rowcolor{tab-lightgrey}
\multirow{-2}{*}{Readability} & sentence\_fre & Flesch-Reading Ease, designed for English & SpaCy, Stanza, TSEval & -0.3976 & 0.19 \\
\multirow{18}{*}{Syntactic} & avg\_distance\_betweeen\_verb\_particle & Average distance between verb and particle based on dependency tree & SpaCy, Stanza, & 0.0385 & \\
 & avg\_distance\_betweeen\_words & Average distance between words based on dependency tree & SpaCy, Stanza, & 0.3934 & 0.17 \\
 & max\_distance\_betweeen\_verb\_particles & Maximum distance between verb and particle based on dependency tree & SpaCy, Stanza, & 0.0385 & \\
 & max\_distance\_betweeen\_words & Maximum distance between words based on dependency tree & SpaCy, Stanza, & 0.2109 & 0.14 \\
 & check\_if\_head\_is\_noun & Whether the head of the dependency tree is a noun & SpaCy, Stanza, TSEval & -0.1147 & 0.13 \\
 & check\_if\_head\_is\_verb & Whether the head of the dependency tree is a verb & SpaCy, Stanza, TSEval & 0.0954 & 0.13 \\
 & check\_if\_one\_child\_of\_root\_is\_subject & Whether a child of a root is a subject (not a verb) & SpaCy, Stanza, TSEval & 0.0144 & 0.15 \\
 & check\_passive\_voice & Whether a sentence is in passive voice & SpaCy, Stanza, TSEval &  &  \\
 & average\_length\_NP & Average length of noun phrase in tokens & SpaCy, Stanza, TSEval & 0.3485 & 0.13 \\
 & average\_length\_VP & Average length of verb phrase in tokens & SpaCy, Stanza, TSEval & 0.4324 & 0.16 \\
 & avg\_length\_PP & Average length of prepositional phrase in tokens & SpaCy, Stanza, TSEval & 0.1406 & 0.07 \\
 & parse\_tree\_height & Depth or height of the dependency tree & SpaCy, Stanza, TSEval & 0.4657 & 0.15 \\
 & ratio\_clauses & Ratio of tokens associated to a clause to all tokens based on dependency tree relations & SpaCy, Stanza, TSEval & 0.1918 & 0.12 \\
 & ratio\_of\_coordinating\_clauses & Ratio of tokens associated to a coordinating clause to all tokens based on dependency tree relations & SpaCy, Stanza, TSEval & 0.2503 & 0.16 \\
 & ratio\_of\_subordinate\_clauses & Ratio of tokens associated to a subordinating clause to all tokens based on dependency tree relations & SpaCy, Stanza, TSEval & 0.2361 & 0.14 \\
 & ratio\_prepositional\_phrases & Ratio of tokens associated to a prepositional phrase to all tokens based on dependency tree relations & SpaCy, Stanza, TSEval & 0.1797 & 0.14 \\
 & ratio\_relative\_phrases & Ratio of tokens associated to a relative clause to all tokens based on dependency tree relations & SpaCy, Stanza, TSEval & 0.3262 & 0.15 \\
 & is\_non\_projective & Whether a dependency tree is non projective & SpaCy, Stanza, TSEval & 0.0883 & 0.13 \\
\rowcolor{tab-lightgrey}
 & ratio\_Abbr\_Yes & Ratio of nouns which are an abbreviation to all nouns & SpaCy, Stanza, UniversalDependencies & -0.0685 & 0.07 \\
\rowcolor{tab-lightgrey}
 & ratio\_Case\_Abe & Ratio of nouns in abessive case to all nouns & SpaCy, Stanza, UniversalDependencies & 0.2394 & \\
\rowcolor{tab-lightgrey}
 & ratio\_Case\_Acc & Ratio of nouns in accusative case to all nouns & SpaCy, Stanza, UniversalDependencies & 0.0658 & 0.29 \\
\rowcolor{tab-lightgrey}
 & ratio\_Case\_Ben & Ratio of nouns in benefactive case to all nouns & SpaCy, Stanza, UniversalDependencies &  &  \\
\rowcolor{tab-lightgrey}
 & ratio\_Case\_Cau & Ratio of nouns in causative case to all nouns & SpaCy, Stanza, UniversalDependencies &  &  \\
\rowcolor{tab-lightgrey}
 & ratio\_Case\_Cmp & Ratio of nouns in comparative case to all nouns & SpaCy, Stanza, UniversalDependencies &  &  \\
\rowcolor{tab-lightgrey}
 & ratio\_Case\_Cns & Ratio of nouns in considerative case to all nouns & SpaCy, Stanza, UniversalDependencies &  &  \\
\rowcolor{tab-lightgrey}
 & ratio\_Case\_Com & Ratio of nouns in comitative case to all nouns & SpaCy, Stanza, UniversalDependencies & 0.1293 & \\
\rowcolor{tab-lightgrey}
 & ratio\_Case\_Dat & Ratio of nouns in dative case to all nouns & SpaCy, Stanza, UniversalDependencies & 0.1822 & 0.09 \\
\rowcolor{tab-lightgrey}
 & ratio\_Case\_Dis & Ratio of nouns in distributive case to all nouns & SpaCy, Stanza, UniversalDependencies &  &  \\
\rowcolor{tab-lightgrey}
 & ratio\_Case\_Equ & Ratio of nouns in equative case to all nouns & SpaCy, Stanza, UniversalDependencies &  &  \\
\rowcolor{tab-lightgrey}
\multirow{-12}{*}{Morphosyntactic} & ratio\_Case\_Erg & Ratio of nouns in ergative case to all nouns & SpaCy, Stanza, UniversalDependencies &  & \\\hline
\end{tabular}
\caption{Overview of all 100 features, including correlation coefficient with CEFR level across all languages.}
\label{tab:allfeature}
\end{sidewaystable*}

\setlength{\tabcolsep}{5pt} 
\begin{sidewaystable*}[!htbp]
\centering
\scriptsize
\begin{tabular}{llllll}
\hline
\textbf{Category} & \textbf{Feature} & \textbf{Short Description} & \textbf{Resource} & \begin{tabular}[c]{@{}l@{}}\textbf{Corr.} $\rho$ \\ \textbf{(avg.)}\end{tabular} & \begin{tabular}[c]{@{}l@{}}\textbf{Corr. }$\rho$ \\ \textbf{(SD)} \end{tabular} \\ \hline
 \multirow{50}{*}{Morphosyntactic} & ratio\_Case\_Ess & Ratio of nouns in essive case to all nouns (relevant for ET) & SpaCy, Stanza, UniversalDependencies & 0.1196 & \\
 & ratio\_Case\_Gen & Ratio of nouns in genitive case to all nouns & SpaCy, Stanza, UniversalDependencies & 0.3528 & 0.12 \\
 & ratio\_Case\_Ins & Ratio of nouns in instrumental case to all nouns (relevant for CZ) & SpaCy, Stanza, UniversalDependencies & 0.2204 & 0.02 \\
 & ratio\_Case\_Nom & Ratio of nouns in nominative case to all nouns & SpaCy, Stanza, UniversalDependencies & -0.0286 & 0.13 \\
 & ratio\_Case\_Par & Ratio of nouns in partitive case to all nouns & SpaCy, Stanza, UniversalDependencies & 0.2252 & \\
 & ratio\_Case\_Tem & Ratio of nouns in temporal case to all nouns & SpaCy, Stanza, UniversalDependencies &  &  \\
 & ratio\_Case\_Tra & Ratio of nouns in translative case to all nouns (relevant for ET) & SpaCy, Stanza, UniversalDependencies & 0.5228 & \\
 & ratio\_Case\_Voc & Ratio of nouns in vocative case to all nouns & SpaCy, Stanza, UniversalDependencies &  &  \\
 & ratio\_Definite\_Com & Ratio of complex nouns to all nouns (relevant for AR) & SpaCy, Stanza, UniversalDependencies &  &  \\
 & ratio\_Definite\_Cons & Ratio of nouns in construct state to all nouns (relevant for AR) & SpaCy, Stanza, UniversalDependencies &  &  \\
 & ratio\_Definite\_Def & Ratio of definite nouns to all nouns & SpaCy, Stanza, UniversalDependencies & 0.0569 & \\
 & ratio\_Definite\_Ind & Ratio of indefinite nouns to all nouns & SpaCy, Stanza, UniversalDependencies & -0.059 & \\
 & ratio\_Foreign\_Yes & Ratio of nouns which are foreign to all nouns & SpaCy, Stanza, UniversalDependencies &  &  \\
 & ratio\_Mood\_Cnd & Ratio of verbs with conditional mood to all verbs & SpaCy, Stanza, UniversalDependencies & 0.2172 & 0.11 \\
 & ratio\_Mood\_Imp & Ratio of verbs with imperative mood to all verbs & SpaCy, Stanza, UniversalDependencies & -0.0243 & 0.1 \\
 & ratio\_Mood\_Ind & Ratio of verbs with indicative mood to all verbs & SpaCy, Stanza, UniversalDependencies & -0.0782 & 0.1 \\
 & ratio\_Mood\_Jus & Ratio of verbs with jussive mood to all verbs & SpaCy, Stanza, UniversalDependencies & 0.0798 & \\
 & ratio\_Mood\_Qot & Ratio of verbs with quotative mood to all verbs & SpaCy, Stanza, UniversalDependencies & 0.1015 & \\
 & ratio\_Mood\_Sub & Ratio of verbs with subjunctive mood to all verbs & SpaCy, Stanza, UniversalDependencies & 0.2076 & 0.13 \\
 & ratio\_Number\_Dual & Ratio of nouns in dual number to all nouns & SpaCy, Stanza, UniversalDependencies & 0.0811 & \\
 & ratio\_Number\_Plur & Ratio of nouns in plural number to all nouns & SpaCy, Stanza, UniversalDependencies & 0.2426 & 0.19 \\
 & ratio\_Number\_Sing & Ratio of nouns in singular number to all nouns & SpaCy, Stanza, UniversalDependencies & 0.0016 & 0.19 \\
 & ratio\_Polarity\_Neg & Ratio of negative verbs to all verbs & SpaCy, Stanza, UniversalDependencies & 0.2287 & 0.19 \\
 & ratio\_Polarity\_Pos & Ratio of positive verbs to all verbs & SpaCy, Stanza, UniversalDependencies & 0.2199 & \\
 & ratio\_Tense\_Fut & Ratio of verbs in future tense to all verbs & SpaCy, Stanza, UniversalDependencies & 0.2023 & 0.07 \\
 & ratio\_Tense\_Imp & Ratio of verbs in imperfect to all verbs & SpaCy, Stanza, UniversalDependencies & 0.2096 & 0.13 \\
 & ratio\_Tense\_Past & Ratio of verbs in past tense to all verbs & SpaCy, Stanza, UniversalDependencies & 0.2782 & 0.17 \\
 & ratio\_Tense\_Pqp & Ratio of verbs in pluferfect to all verbs & SpaCy, Stanza, UniversalDependencies &  &  \\
 & ratio\_Tense\_Pres & Ratio of verbs in present tenst to all verbs & SpaCy, Stanza, UniversalDependencies & -0.1113 & 0.16 \\
 & ratio\_Voice\_Act & Ratio of verbs in active voice to all verbs & SpaCy, Stanza, UniversalDependencies & 0.0225 & 0.18 \\
 & ratio\_Voice\_Mid & Ratio of verbs in middle voice to all verbs & SpaCy, Stanza, UniversalDependencies & 0.0682 & \\
 & ratio\_Voice\_Pass & Ratio of verbs in passive voice to all verbs & SpaCy, Stanza, UniversalDependencies & 0.3196 & 0.17 \\
 & ratio\_mwes & Ratio of multi-word expressions to all tokens based on dependency tree relations & SpaCy, Stanza, TSEval & 0.1171 & 0.13 \\
 & ratio\_named\_entities & Ratio of multi-word expressions to all tokens based on SpaCy pipeline & SpaCy, Stanza, TSEval &  &  \\
 & ratio\_of\_adjectives & Ratio of adjectives to all tokens & SpaCy, Stanza, TSEval & 0.172 & 0.14 \\
 & ratio\_of\_adpositions & Ratio of adpositions to all tokens & SpaCy, Stanza, TSEval & 0.2342 & 0.13 \\
 & ratio\_of\_adverbs & Ratio of adverbs to all tokens & SpaCy, Stanza, TSEval & 0.07 & 0.19 \\
 & ratio\_of\_auxiliary\_verbs & Ratio of auxiliary verbs to all tokens & SpaCy, Stanza, TSEval & 0.0034 & 0.13 \\
 & ratio\_of\_conjunctions & Ratio of conjunctions to all tokens & SpaCy, Stanza, TSEval & 0.1429 & 0.13 \\
 & ratio\_of\_determiners & Ratio of determiners to all tokens & SpaCy, Stanza, TSEval & 0.2459 & 0.16 \\
 & ratio\_of\_function\_words & Ratio of function words to all tokens based on dependency tree relations & SpaCy, Stanza, TSEval & 0.238 & 0.16 \\
 & ratio\_of\_interjections & Ratio of interjections to all tokens & SpaCy, Stanza, TSEval & -0.201 & 0.15 \\
 & ratio\_of\_nouns & Ratio of nouns to all tokens & SpaCy, Stanza, TSEval & -0.0509 & 0.17 \\
 & ratio\_of\_numerals & Ratio of numerals to all tokens & SpaCy, Stanza, TSEval & -0.005 & 0.21 \\
 & ratio\_of\_particles & Ratio of particles to all tokens & SpaCy, Stanza, TSEval & 0.143 & 0.16 \\
 & ratio\_of\_pronouns & Ratio of pronouns to all tokens & SpaCy, Stanza, TSEval & -0.0809 & 0.14 \\
 & ratio\_of\_punctuation & Ratio of punctuation marks to all tokens & SpaCy, Stanza, TSEval & -0.2226 & 0.16 \\
 & ratio\_of\_symbols & Ratio of symbols to all tokens & SpaCy, Stanza, TSEval & 0.0542 & 0.11 \\
 & ratio\_of\_verbs & Ratio of verbs to all tokens & SpaCy, Stanza, TSEval & -0.0954 & 0.14 \\
 & verb\_noun\_ratio & How many verbs occur per noun? The higher the value (the more verbs), the easier the text & SpaCy, Stanza, TSEval & 0.0501 & 0.21 \\
 \hline
\end{tabular}
\caption{Overview of all 100 features, including correlation coefficient with CEFR level across all languages. Part II.}
\label{tab:allfeature-2}
\end{sidewaystable*}



\section{Additional Context on Restrictions of GDPR-Protected Datasets}
\label{app:additional_context_gdpr}
The critical aspect of the GDPR is that it gives data subjects (e.g., L2 learners of CEFR) the right to withdraw their personal information from processing, which requires data processors to store both the signed consents and the ID mappings (i.e., mappings between the names of the real people and their IDs in a released corpora). As long as these documents exist and reidentification is theoretically possible, the data falls under the scope of the GDPR. Further complicating factors are national legislations and ethical regulations, such as archival laws, that treat any data produced at universities---including those used for language proficiency assessment such as essays, recorded dialogues, and written texts from personal experiences---as the property of the state (and hence making destruction of the ID mappings a non-trivial act) \cite{gdpr2016}.

Yet another upcoming challenge is the EU AI Act \cite{aiact2024} that implies that AI models trained on personal data should inherit the same license as the data they have been trained on, meaning that the models will be under the scope of the GDPR. We hypothesize that the non-restricted datasets included in \textsc{UniversalCEFR} either do not contain personal information or were collected before the GDPR, since they are already openly accessible to the public. We further hypothesize that the datasets currently under the GDPR will eventually have their ID mappings destroyed and will no longer be subject to the GDPR. This may mean that the learner corpora that can be added to \textsc{UniversalCEFR} will grow with time.

\section{Full Dataset Directory of UniversalCEFR}
\label{app:full_dataset_director}

We provide the complete information of qualified corpora included in the current \textsc{UniversalCEFR} collection to form a directory of datasets. Aside from eight per-instance information included in the standardized JSON format in Table~\ref{tab:json_fields}, we also report five per-corpus information as listed below:

\begin{itemize}
    \item Annotation method used (manual, computer-assisted, or NA).
    \item Total number of expert annotators.
    \item Distinct L1 learners per language for learner corpora.
    \item Inter-annotator agreement (IAA) metric and score.
    \item Reference to published paper or repository.
\end{itemize}

\section{Prompt Templates}
\label{app:prompt_templates}

We provide the complete copies of the prompt templates used in prompting experiments with instruction-tuned LLMs as described in Section~\ref{sec:cefr_level_classification}. The prompt templates are categorized by color based on the setup: \textcolor{gray}{\textbf{\textsc{Base}}}, \textcolor{teal}{\textbf{\textsc{En-Read}}}, \textcolor{teal}{\textbf{\textsc{Lang-Read}}},  \textcolor{violet}{\textbf{\textsc{En-Write}}},  \textcolor{violet}{\textbf{\textsc{Lang-Write}}}.

\section{Welsh Data Collection}
\label{app:welsh_data_collection}
One of the contributions of \textsc{UniversalCEFR} is the release of the first-ever open dataset for the Welsh language (\textsc{CY}) with gold-standard CEFR labels for A1 and A2. To obtain this data, we corresponded with data maintainers from Learn Welsh (\url{https://learnwelsh.cymru/}), which is a compilation of expert-created books (reference texts) and acquired PDF versions. This resource can be shared in any format for non-commercial research, which fits the goal of \textsc{UniversalCEFR}. We then manually extracted qualified texts according to the four levels of granularity: sentence, paragraph, dialogue, and document. The distribution of CEFR levels and text granularity for this new Welsh dataset can be found in Table~\ref{tab:universalcefr_main_count_grade_coverage} and \ref{tab:universalcefr_main_count_level}, respectively.

\setlength{\tabcolsep}{5pt} 
\begin{sidewaystable*}[!htbp]
\scriptsize
\begin{tabular}{@{}lllllllllllll@{}}
\toprule
Corpus Name &
  \begin{tabular}[c]{@{}l@{}}Lang Code\\(ISO 638-1)\end{tabular} &
  Format &
  Category &
  Size &
  \begin{tabular}[c]{@{}l@{}}Annotation\\ Method\end{tabular} &
  \begin{tabular}[c]{@{}l@{}}Expert\\ Annotators\end{tabular} &
  Distinct L1 &
  \begin{tabular}[c]{@{}l@{}}Inter-Annotator\\ Agreement\end{tabular}  &
  CEFR Coverage &
  License &
  Resource \\ \midrule
\texttt{cambridge-exams} &
  en &
  document-level &
  reference &
  331 &
  n/a &
  n/a &
  n/a &
  n/a &
  A1-C2 &
  CC BY-NC-SA 4.0 &
  \citet{xia-etal-2016-text} \\
\texttt{elg-cefr-en} &
  en &
  document-level &
  reference &
  712 &
  manual &
  3 &
  n/a &
  n/a &
  A1-C2, plus &
  CC BY-NC-SA 4.0 &
  \citet{breuker2022cefr} \\
\texttt{cefr-sp} &
  en &
  sentence-level &
  reference &
  17,000 &
  manual &
  2 &
  n/a &
  $r = 0.75, 0.73$  &
  A1-C2 &
  CC BY-NC-SA 4.0 &
  \citet{arase-etal-2022-cefr} \\
\texttt{elg-cefr-de} &
  de &
  document-level &
  reference &
  509 &
  manual &
  3 &
  n/a &
  n/a &
  A1-C2 &
  CC BY-NC-SA 4.0 &
  \citet{breuker2022cefr} \\
\texttt{elg-cefr-nl} &
  nl &
  document-level &
  reference &
  3,596 &
  manual &
  3 &
  n/a &
  n/a &
  A1-C2, plus &
  CC BY-NC-SA 4.0 &
  \citet{breuker2022cefr} \\
\texttt{icle500} &
  en &
  document-level &
  learner &
  500 &
  manual &
  28 &
  \begin{tabular}[c]{@{}l@{}}ur, pa, bg, zh, cs, \\ nl, fi, fr, de, el, hu, \\ it, ja, ko, lt, mk, no, \\ fa, pl, pt, ru, sr, es, \\ sv, tn, tr\end{tabular} &
  Rasch $\kappa = -0.02$ &
  A1-C2, plus &
  CC BY-NC 4.0 &
  \begin{tabular}[c]{@{}l@{}}\citet{thwaites2024crowdsourced}, \\ \citet{granger2009international} \end{tabular} \\
\texttt{cefr-asag} &
  en &
  paragraph-level &
  learner &
  299 &
  manual &
  3 &
  fr &
  Krippendorf $\alpha = 0.81$ &
  A1-C2 &
  CC BY-NC-SA 4.0 &
  \citet{tack-etal-2017-human} \\
\texttt{merlin-cs} &
  cs &
  paragraph-level &
  learner &
  441 &
  manual &
  multiple &
  \begin{tabular}[c]{@{}l@{}}hu, de, fr, ru, pl, \\ en, sk, es\end{tabular} &
  n/a &
  A2-B2 &
  CC BY-SA 4.0 &
  \citet{boyd-etal-2014-merlin} \\
\texttt{merlin-it} &
  it &
  paragraph-level &
  learner &
  813 &
  manual &
  multiple &
  \begin{tabular}[c]{@{}l@{}}hu, de, fr, ru, pl, \\ en, sk, es\end{tabular} &
  n/a &
  A1-B1 &
  CC BY-SA 4.0 &
  \citet{boyd-etal-2014-merlin} \\
\texttt{merlin-de} &
  de &
  paragraph-level &
  learner &
  1,033 &
  manual &
  multiple &
  \begin{tabular}[c]{@{}l@{}}hu, de, fr, ru, pl, \\ en, sk, es\end{tabular} &
  n/a &
  A1-C1 &
  CC BY-SA 4.0 &
  \citet{boyd-etal-2014-merlin} \\
\texttt{hablacultura} &
  es &
  paragraph-level &
  reference &
  710 &
  manual &
  multiple &
  n/a &
  n/a &
  A2-C1 &
  CC BY NC 4.0 &
  \citet{vasquez-rodriguez-etal-2022-benchmark} \\
\texttt{kwiziq-es} &
  es &
  document-level &
  reference &
  206 &
  manual &
  multiple &
  n/a &
  n/a &
  A1-C1 &
  CC BY NC 4.0 &
  \citet{vasquez-rodriguez-etal-2022-benchmark} \\
\texttt{kwiziq-fr} &
  fr &
  document-level &
  reference &
  344 &
  manual &
  multiple &
  n/a &
  n/a &
  A1-C1 &
  CC BY NC 4.0 &
  Original \\
\texttt{caes} &
  es &
  document-level &
  learner &
  30,935 &
  computer-assisted &
  multiple &
  pt, zh, ar, fr, ru &
  n/a &
  A1-C1 &
  CC BY NC 4.0 &
  \citet{vasquez-rodriguez-etal-2022-benchmark} \\
\texttt{deplain-web-doc} &
  de &
  document-level &
  reference &
  394 &
  manual &
  2 &
  n/a &
  Cohen $\kappa = 0.85$ &
  A1,A2,B2,C2 &
  \begin{tabular}[c]{@{}l@{}}CC-BY-SA-3, \\ CC-BY-4, \\ CC-BY-NC-ND-4, \\ save\_use\_share\end{tabular} &
  \citet{stodden-etal-2023-deplain} \\
\texttt{deplain-apa-doc} &
  de &
  document-level &
  reference &
  483 &
  manual &
  2 &
  n/a &
  Cohen $\kappa = 0.85$ &
  A2-B1 &
  \begin{tabular}[c]{@{}l@{}}CC-BY-SA-3, \\ CC-BY-4, \\ CC-BY-NC-ND-4, \\ save\_use\_share\end{tabular} &
  \citet{stodden-etal-2023-deplain} \\
\texttt{deplain-apa-sent} &
  de &
  sentence-level &
  reference &
  483 &
  manual &
  2 &
  n/a &
  n/a &
  A2-B2 &
  By request &
  \citet{stodden-etal-2023-deplain} \\
\texttt{elle} &
  et &
  \begin{tabular}[c]{@{}l@{}}paragraph-level, \\ document-level\end{tabular} &
  learner &
  1,697 &
  manual &
  2 &
  n/a &
  n/a &
  A2-C1 &
  CC BY 4.0 &
  \begin{tabular}[c]{@{}l@{}}\citet{allkivi2024elle}, \\ \citet{vajjala-rama-2018-experiments} \end{tabular} \\
\texttt{efcamdat-cleaned} &
  en &
  \begin{tabular}[c]{@{}l@{}}sentence-level,\\ paragraph-level\end{tabular} &
  learner &
  406,062 &
  manual &
  n/a &
  \begin{tabular}[c]{@{}l@{}}br, zh, tw, ru, sa, \\ mx, de, it, fr, jp, tr\end{tabular} &
  n/a &
  A1-C1 &
  Cambridge &
  \begin{tabular}[c]{@{}l@{}}\citet{geertzen2013automatic}, \\ \citet{shatz2020refining} \\ \citet{huang2017ef} \end{tabular} \\
\texttt{beast2019-w\&i} &
  en &
  sentence-level &
  learner &
  3,600 &
  manual &
  multiple &
  n/a &
  n/a &
  A1-C2 &
  Cambridge &
  \begin{tabular}[c]{@{}l@{}}\citet{bryant-etal-2019-bea}, \\ \citet{yannakoudakis2018developing} \end{tabular} \\
\texttt{peapl2} &
  pt &
  paragraph-level &
  learner &
  481 &
  manual &
  n/a &
  \begin{tabular}[c]{@{}l@{}}zh, en, es, de, ru, \\ fr, ja, it, nl, tet, ar, \\ pl, ko, ro, sv\end{tabular} &
  n/a &
  A1-C2 &
  CC BY SA NC 4.0 &
  \citet{martins2019corpus} \\
\texttt{cople2} &
  pt &
  paragraph-level &
  learner &
  942 &
  manual &
  n/a &
  \begin{tabular}[c]{@{}l@{}}zh, en, es, de, ru, \\ fr, ja, it, nl, tet, ar, \\ pl, ko, ro, sv\end{tabular} &
  n/a &
  A1-C1 &
  CC BY SA NC 4.0 &
  \citet{mendes-etal-2016-cople2} \\
\texttt{zaebuc} &
  ar &
  paragraph-level &
  learner &
  214 &
  manual &
  3 &
  en &
  Unnamed $\kappa = 0.99$ &
  A2-C1 &
  CC BY SA NC 4.0 &
  \citet{habash-palfreyman-2022-zaebuc} \\
\texttt{readme} &
  \begin{tabular}[c]{@{}l@{}}ar, en, fr, \\ hi, ru\end{tabular} &
  sentence-level &
  reference &
  9,757 &
  computer-assisted &
  2 &
  n/a &
  Krippendorf $\kappa = 0.67,0.78$ &
  A1-C2 &
  CC BY SA NC 4.0 &
  \citet{naous-etal-2024-readme} \\
\texttt{apa-lha} &
  de &
  document-level &
  reference &
  3,130 &
  n/a &
  n/a &
  n/a &
  n/a &
  A2-B1 &
  Public &
  \citet{spring-etal-2021-exploring} \\
\texttt{learn-welsh} &
  cy &
  \begin{tabular}[c]{@{}l@{}}document-level, \\ sentence-level,\\ discourse-level\end{tabular} &
  reference &
  1,372 &
  manual &
  n/a &
  n/a &
  n/a &
  A1-A2 &
  Public &
  Original \\ 
  \bottomrule
\end{tabular}
\caption{The \textsc{UniversalCEFR-Full} directory of dataset information reporting full details of properties of corpora included in the main collection.}
\label{tab:universalcefr_full_table}
\end{sidewaystable*}

\clearpage

\onecolumn

    \begin{tcolorbox}[colframe=gray, colback=white, title=Base CEFR prompt template, coltitle=white, center title, fonttitle=\bfseries]

    You are an expert in language proficiency classification based on the Common European Framework of Reference for Languages (CEFR). Your task is to analyze the given text or narrative and determine its CEFR level [A1, A2, B1, B2, C1, or C2] based on vocabulary complexity, grammar, and overall language proficiency. Provide only the CEFR level as output, without explanation or justification.\\
    
    Text: <<\texttt{TEXT}>> \\
    
    Answer:

    \end{tcolorbox}


    \begin{tcolorbox}[colframe=teal, colback=white, title=CEFR specifications for reading comprehension in English (\textsc{en}), coltitle=white, center title, fonttitle=\bfseries]

    You are an expert in language proficiency classification based on the Common European Framework of Reference for Languages (CEFR). Your task is to analyze the given text or narrative and determine the best CEFR level [A1, A2, B1, B2, C1, or C2] based on the CEFR descriptors of reading comprehension of learners below:\\

    A1 - Learners of this level can understand very short, simple texts a single phrase at a time, picking up familiar names, words and basic phrases and rereading as required.\\
    
    A2 - Learners of this level can understand short, simple texts containing the highest frequency vocabulary, including a proportion of shared international vocabulary items.\\
    
    B1 - Learners of this level can read straightforward factual texts on subjects related to their field of interest with a satisfactory level of comprehension.\\
    
    B2 - Learners of this level can read with a large degree of independence, adapting style and speed of reading to different texts and purposes, and using appropriate reference sources selectively. Has a broad active reading vocabulary, but may experience some difficulty with low-frequency idioms.\\
    
    C1 - Learners of this level can understand in detail lengthy, complex texts, whether or not these relate to their own area of speciality, provided they can reread difficult sections. They can also understand a wide variety of texts including literary writings, newspaper or magazine articles, and specialized academic or professional publications, provided there are opportunities for rereading and they have access to reference tools.\\
    
    C2 - Learners of this level can understand virtually all types of texts including abstract, structurally complex, or highly colloquial literary and non-literary writings. They can also understand a wide range of long and complex texts, appreciating subtle distinctions of style and implicit as well as explicit meaning.\\
    
    Provide only the CEFR level as output directly, without explanation or justification.\\
    
    Text: <<\texttt{TEXT}>> \\
    
    Answer:

    \end{tcolorbox}

    \begin{tcolorbox}[colframe=teal, colback=white, title=CEFR specifications for reading comprehension in Spanish (\textsc{es}), coltitle=white, center title, fonttitle=\bfseries]

    You are an expert in language proficiency classification based on the Common European Framework of Reference for Languages (CEFR). Your task is to analyze the given \textbf{Spanish} text or narrative and determine the best CEFR level [A1, A2, B1, B2, C1, or C2] based on the CEFR descriptors of reading comprehension of learners below:\\
    
    A1 - Los estudiantes de este nivel pueden comprender textos muy breves y sencillos, frase por frase, recogiendo nombres, palabras y frases básicas familiares y releyendo según sea necesario.\\
    
    A2 - Los estudiantes de este nivel pueden comprender textos breves y sencillos que contienen el vocabulario de mayor frecuencia, incluyendo una proporción de vocabulario internacional compartido.\\
    
    B1 - Los estudiantes de este nivel pueden leer textos factuales sencillos sobre temas relacionados con su área de interés con un nivel de comprensión satisfactorio.\\
    
    B2 - Los estudiantes de este nivel pueden leer con un alto grado de independencia, adaptando el estilo y la velocidad de lectura a diferentes textos y propósitos, y utilizando selectivamente las fuentes de referencia adecuadas. Poseen un amplio vocabulario de lectura activa, pero pueden tener alguna dificultad con expresiones idiomáticas de baja frecuencia.\\
    
    C1 - Los estudiantes de este nivel pueden comprender con detalle textos extensos y complejos, independientemente de si se relacionan con su área de especialidad, siempre que puedan releer las secciones difíciles. También pueden comprender una amplia variedad de textos, incluyendo escritos literarios, artículos de periódicos o revistas, y publicaciones académicas o profesionales especializadas, siempre que tengan la oportunidad de releer y acceso a recursos de referencia.\\
    
    C2 - Los estudiantes de este nivel pueden comprender prácticamente todo tipo de textos, incluyendo textos literarios y no literarios abstractos, estructuralmente complejos o muy coloquiales. También pueden comprender una amplia gama de textos largos y complejos, apreciando las sutiles diferencias de estilo y el significado, tanto implícito como explícito.\\
    
    Provide only the CEFR level as output directly, without explanation or justification.\\
    
    Text: <<\texttt{TEXT}>> \\
    
    Answer:

    \end{tcolorbox}

    \begin{tcolorbox}[colframe=teal, colback=white, title=CEFR specifications for reading comprehension in German (\textsc{de}), coltitle=white, center title, fonttitle=\bfseries]

    You are an expert in language proficiency classification based on the Common European Framework of Reference for Languages (CEFR). Your task is to analyze the given \textbf{German} text or narrative and determine the best CEFR level [A1, A2, B1, B2, C1, or C2] based on the CEFR descriptors of reading comprehension of learners below:\\
    
    A1 – Lernende dieser Stufe können sehr kurze, einfache Texte Satz für Satz verstehen, indem sie bekannte Namen, Wörter und einfache Sätze aufgreifen und bei Bedarf wiederholt lesen.\\
    
    A2 – Lernende dieser Stufe können kurze, einfache Texte mit dem häufigsten Wortschatz verstehen, darunter auch einen Anteil an international verbreiteten Vokabeln.\\
    
    B1 – Lernende dieser Stufe können einfache Sachtexte zu Themen ihres Interessengebiets mit zufriedenstellendem Verständnis lesen.\\
    
    B2 – Lernende dieser Stufe können weitgehend selbstständig lesen, indem sie Stil und Geschwindigkeit an unterschiedliche Texte und Zwecke anpassen und geeignete Referenzquellen selektiv nutzen. Sie verfügen über einen breiten aktiven Lesewortschatz, haben aber möglicherweise Schwierigkeiten mit seltenen Redewendungen.\\
    
    C1 – Lernende dieser Stufe können längere, komplexe Texte detailliert verstehen, unabhängig davon, ob sie zu ihrem Fachgebiet gehören oder nicht, sofern sie schwierige Abschnitte wiederholt lesen können. Sie können außerdem eine Vielzahl von Texten verstehen, darunter literarische Schriften, Zeitungs- und Zeitschriftenartikel sowie wissenschaftliche oder professionelle Fachpublikationen, sofern Möglichkeiten zum Nachlesen bestehen und sie Zugang zu Nachschlagewerken haben.\\
    
    C2 – Lernende dieser Stufe können nahezu alle Textarten verstehen, darunter abstrakte, strukturell komplexe oder stark umgangssprachliche literarische und nicht-literarische Texte. Sie können außerdem eine breite Palette langer und komplexer Texte verstehen und dabei subtile Stilunterschiede sowie implizite und explizite Bedeutungen wahrnehmen.\\
    
    Provide only the CEFR level as output directly, without explanation or justification.\\
    
    Text: <<\texttt{TEXT}>> \\
    
    Answer:

    \end{tcolorbox}

    \begin{tcolorbox}[colframe=teal, colback=white, title=CEFR specifications for reading comprehension in Dutch (\textsc{nl}), coltitle=white, center title, fonttitle=\bfseries]

    You are an expert in language proficiency classification based on the Common European Framework of Reference for Languages (CEFR). Your task is to analyze the given \textbf{Dutch} text or narrative and determine the best CEFR level [A1, A2, B1, B2, C1, or C2] based on the CEFR descriptors of reading comprehension of learners below:\\
    
    A1 - Leerlingen van dit niveau kunnen zeer korte, eenvoudige teksten begrijpen, één zin tegelijk, bekende namen, woorden en basiszinnen oppikken en indien nodig herlezen.\\
    
    A2 - Leerlingen van dit niveau kunnen korte, eenvoudige teksten begrijpen die de meest frequente woordenschat bevatten, inclusief een deel van de gedeelde internationale woordenschatitems.\\
    
    B1 - Leerlingen van dit niveau kunnen eenvoudige feitelijke teksten lezen over onderwerpen die verband houden met hun interessegebied met een bevredigend niveau van begrip.\\
    
    B2 - Leerlingen van dit niveau kunnen met een grote mate van onafhankelijkheid lezen, de stijl en leessnelheid aanpassen aan verschillende teksten en doeleinden, en selectief gebruikmaken van geschikte referentiebronnen. Heeft een brede actieve leeswoordenschat, maar kan enige moeite hebben met laagfrequente idiomen.\\
    
    C1 - Leerlingen van dit niveau kunnen lange, complexe teksten gedetailleerd begrijpen, ongeacht of deze betrekking hebben op hun eigen vakgebied, op voorwaarde dat ze moeilijke secties kunnen herlezen. Ze kunnen ook een breed scala aan teksten begrijpen, waaronder literaire geschriften, kranten- of tijdschriftartikelen en gespecialiseerde academische of professionele publicaties, mits er mogelijkheden zijn om ze opnieuw te lezen en ze toegang hebben tot referentietools.\\
    
    C2 - Cursisten van dit niveau kunnen vrijwel alle soorten teksten begrijpen, waaronder abstracte, structureel complexe of zeer informele literaire en niet-literaire geschriften. Ze kunnen ook een breed scala aan lange en complexe teksten begrijpen, waarbij ze subtiele verschillen in stijl en impliciete en expliciete betekenis waarderen.\\
    
    Provide only the CEFR level as output directly, without explanation or justification.\\
    
    Text: <<\texttt{TEXT}>> \\
    
    Answer:

    \end{tcolorbox}

    \begin{tcolorbox}[colframe=teal, colback=white, title=CEFR specifications for reading comprehension in Czech (\textsc{cs}), coltitle=white, center title, fonttitle=\bfseries]

    You are an expert in language proficiency classification based on the Common European Framework of Reference for Languages (CEFR). Your task is to analyze the given \textbf{Czech} text or narrative and determine the best CEFR level [A1, A2, B1, B2, C1, or C2] based on the CEFR descriptors of reading comprehension of learners below:\\
    
    A1 – Studenti této úrovně dokážou porozumět velmi krátkým jednoduchým textům po jedné frázi, pochytají známá jména, slova a základní fráze a přečtou si je podle potřeby.\\
    
    A2 – Studenti této úrovně dokážou porozumět krátkým jednoduchým textům obsahujícím nejfrekventovanější slovní zásobu, včetně části sdílené mezinárodní slovní zásoby.\\
    
    B1 – Studenti této úrovně dokážou číst přímočaré věcné texty na témata související s jejich oblastí zájmu s uspokojivou úrovní porozumění.\\
    
    B2 – Studenti této úrovně dokážou číst s velkou mírou nezávislosti, přizpůsobují styl a rychlost čtení různým textům a účelům a selektivně používají vhodné referenční zdroje. Má širokou slovní zásobu aktivního čtení, ale může mít potíže s nízkofrekvenčními idiomy.\\
    
    C1 – Studenti této úrovně dokážou podrobně porozumět dlouhým a složitým textům, ať už se týkají nebo netýkají jejich vlastní oblasti specializace, za předpokladu, že dokážou znovu přečíst obtížné části. Mohou také porozumět široké škále textů, včetně literárních textů, článků v novinách nebo časopisech a specializovaných akademických nebo odborných publikací, za předpokladu, že mají příležitosti k opakovanému čtení a mají přístup k referenčním nástrojům.\\
    
    C2 – Studenti této úrovně mohou porozumět prakticky všem typům textů včetně abstraktních, strukturálně složitých nebo vysoce hovorových literárních a neliterárních spisů. Dokážou také porozumět široké škále dlouhých a složitých textů, ocenit jemné rozdíly ve stylu a implicitní i explicitní význam.\\
    
    Provide only the CEFR level as output directly, without explanation or justification.\\
    
    Text: <<\texttt{TEXT}>> \\
    
    Answer:

    \end{tcolorbox}

    \begin{tcolorbox}[colframe=teal, colback=white, title=CEFR specifications for reading comprehension in Italian (\textsc{it}), coltitle=white, center title, fonttitle=\bfseries]

    You are an expert in language proficiency classification based on the Common European Framework of Reference for Languages (CEFR). Your task is to analyze the given \textbf{Italian} text or narrative and determine the best CEFR level [A1, A2, B1, B2, C1, or C2] based on the CEFR descriptors of reading comprehension of learners below:\\
    
    A1 - Gli studenti di questo livello riescono a comprendere testi molto brevi e semplici, una frase alla volta, cogliendo nomi familiari, parole e frasi di base e rileggendo quando necessario.\\
    
    A2 - Gli studenti di questo livello riescono a comprendere testi brevi e semplici contenenti il vocabolario più frequente, inclusa una parte di elementi di vocabolario internazionale condiviso.\\
    
    B1 - Gli studenti di questo livello riescono a leggere testi fattuali semplici su argomenti correlati al loro campo di interesse con un livello di comprensione soddisfacente.\\
    
    B2 - Gli studenti di questo livello riescono a leggere con un ampio grado di indipendenza, adattando stile e velocità di lettura a testi e scopi diversi e utilizzando fonti di riferimento appropriate in modo selettivo. Ha un ampio vocabolario di lettura attiva, ma può avere qualche difficoltà con idiomi a bassa frequenza.\\
    
    C1 - Gli studenti di questo livello riescono a comprendere in dettaglio testi lunghi e complessi, indipendentemente dal fatto che siano correlati o meno alla propria area di specializzazione, a condizione che riescano a rileggere sezioni difficili. Possono anche comprendere un'ampia varietà di testi, tra cui scritti letterari, articoli di giornali o riviste e pubblicazioni accademiche o professionali specializzate, a condizione che vi siano opportunità di rilettura e abbiano accesso a strumenti di riferimento.\\
    
    C2 - Gli studenti di questo livello possono comprendere praticamente tutti i tipi di testi, tra cui scritti letterari e non letterari astratti, strutturalmente complessi o altamente colloquiali. Possono anche comprendere un'ampia gamma di testi lunghi e complessi, apprezzando sottili distinzioni di stile e significato implicito ed esplicito.\\

    Provide only the CEFR level as output directly, without explanation or justification.\\
    
    Text: <<\texttt{TEXT}>> \\
    
    Answer:

    \end{tcolorbox}

    \begin{tcolorbox}[colframe=teal, colback=white, title=CEFR specifications for reading comprehension in French (\textsc{fr}), coltitle=white, center title, fonttitle=\bfseries]

    You are an expert in language proficiency classification based on the Common European Framework of Reference for Languages (CEFR). Your task is to analyze the given \textbf{French} text or narrative and determine the best CEFR level [A1, A2, B1, B2, C1, or C2] based on the CEFR descriptors of reading comprehension of learners below:\\
    
    A1 - Les apprenants de ce niveau peuvent comprendre des textes très courts et simples, phrase par phrase, en reprenant des noms, des mots et des expressions de base familiers et en les relisant si nécessaire.\\
    
    A2 - Les apprenants de ce niveau peuvent comprendre des textes courts et simples contenant le vocabulaire le plus courant, y compris une partie du vocabulaire international commun.\\
    
    B1 - Les apprenants de ce niveau peuvent lire des textes factuels simples sur des sujets liés à leur domaine d'intérêt avec un niveau de compréhension satisfaisant.\\
    
    B2 - Les apprenants de ce niveau peuvent lire avec une grande autonomie, en adaptant leur style et leur vitesse de lecture à différents textes et objectifs, et en utilisant sélectivement des sources de référence appropriées. Possède un vocabulaire de lecture actif et étendu, mais peut éprouver des difficultés avec les expressions idiomatiques peu fréquentes.\\
    
    C1 - Les apprenants de ce niveau peuvent comprendre en détail des textes longs et complexes, qu'ils relèvent ou non de leur domaine de spécialité, à condition de pouvoir relire les passages difficiles. Ils peuvent également comprendre une grande variété de textes, notamment des écrits littéraires, des articles de journaux ou de magazines, ainsi que des publications universitaires ou professionnelles spécialisées, à condition de disposer d'opportunités de relecture et d'outils de référence.\\
    
    C2 - Les apprenants de ce niveau peuvent comprendre pratiquement tous les types de textes, y compris les écrits littéraires et non littéraires abstraits, structurellement complexes ou très familiers. Ils peuvent également comprendre un large éventail de textes longs et complexes, en appréciant les subtilités stylistiques et le sens implicite et explicite.\\

    Provide only the CEFR level as output directly, without explanation or justification.\\
    
    Text: <<\texttt{TEXT}>> \\
    
    Answer:

    \end{tcolorbox}

    \begin{tcolorbox}[colframe=teal, colback=white, title=CEFR specifications for reading comprehension in Estonian (\textsc{et}), coltitle=white, center title, fonttitle=\bfseries]

    You are an expert in language proficiency classification based on the Common European Framework of Reference for Languages (CEFR). Your task is to analyze the given \textbf{Estonian} text or narrative and determine the best CEFR level [A1, A2, B1, B2, C1, or C2] based on the CEFR descriptors of reading comprehension of learners below:\\

    A1 – selle taseme õppijad saavad aru väga lühikestest lihtsatest tekstidest ühe fraasi kaupa, korjavad üles tuttavad nimed, sõnad ja põhifraasid ning loevad vajaduse korral uuesti läbi.\\
    
    A2 – selle taseme õppijad saavad aru lühikestest lihtsatest tekstidest, mis sisaldavad kõige sagedamini kasutatavat sõnavara, sealhulgas osa jagatud rahvusvahelistest sõnavaraüksustest.\\
    
    B1 – selle taseme õppijad oskavad rahuldaval mõistustasemel lugeda otsekoheseid faktitekste nende huvivaldkonnaga seotud teemadel.\\
    
    B2 – selle taseme õppijad oskavad lugeda suurel määral iseseisvalt, kohandades lugemisstiili ja -kiirust erinevate tekstide ja eesmärkidega ning kasutades valikuliselt sobivaid viiteallikaid. Tal on lai aktiivse lugemise sõnavara, kuid tal võib esineda raskusi madala sagedusega idioomidega.\\
    
    C1 – selle taseme õppijad saavad üksikasjalikult aru pikkadest ja keerukatest tekstidest, olenemata sellest, kas need on seotud nende enda erialaga või mitte, eeldusel, et nad suudavad raskeid lõike uuesti lugeda. Nad saavad aru ka paljudest erinevatest tekstidest, sealhulgas kirjanduslikest kirjutistest, ajalehtede või ajakirjade artiklitest ning erialastest akadeemilistest või erialastest väljaannetest, eeldusel, et neil on võimalus uuesti lugeda ja neil on juurdepääs viitevahenditele.\\
    
    C2 – selle taseme õppijad saavad aru peaaegu igat tüüpi tekstidest, sealhulgas abstraktsetest, struktuurselt keerukatest või väga kõnekeelsetest kirjanduslikest ja mittekirjanduslikest kirjutistest. Samuti saavad nad aru paljudest pikkadest ja keerulistest tekstidest, mõistes peent stiilieritlust ning kaudset ja selgesõnalist tähendust.\\

    Provide only the CEFR level as output directly, without explanation or justification.\\
    
    Text: <<\texttt{TEXT}>> \\
    
    Answer:

    \end{tcolorbox}

    \begin{tcolorbox}[colframe=teal, colback=white, title=CEFR specifications for reading comprehension in Portuguese (\textsc{pt}), coltitle=white, center title, fonttitle=\bfseries]

    You are an expert in language proficiency classification based on the Common European Framework of Reference for Languages (CEFR). Your task is to analyze the given \textbf{Portuguese} text or narrative and determine the best CEFR level [A1, A2, B1, B2, C1, or C2] based on the CEFR descriptors of reading comprehension of learners below:\\
    
    A1 - Os alunos deste nível podem entender textos muito curtos e simples, uma única frase de cada vez, pegando nomes, palavras e frases básicas familiares e relendo conforme necessário.\\
    
    A2 - Os alunos deste nível podem entender textos curtos e simples contendo o vocabulário de maior frequência, incluindo uma proporção de itens de vocabulário internacional compartilhados.\\
    
    B1 - Os alunos deste nível podem ler textos factuais diretos sobre assuntos relacionados ao seu campo de interesse com um nível satisfatório de compreensão.\\
    
    B2 - Os alunos deste nível podem ler com um alto grau de independência, adaptando o estilo e a velocidade de leitura a diferentes textos e propósitos, e usando fontes de referência apropriadas seletivamente. Tem um amplo vocabulário de leitura ativa, mas pode ter alguma dificuldade com expressões idiomáticas de baixa frequência.\\
    
    C1 - Os alunos deste nível podem entender em detalhes textos longos e complexos, estejam eles relacionados ou não à sua própria área de especialidade, desde que possam reler seções difíceis. Eles também podem entender uma grande variedade de textos, incluindo escritos literários, artigos de jornais ou revistas e publicações acadêmicas ou profissionais especializadas, desde que haja oportunidades de releitura e tenham acesso a ferramentas de referência.\\
    
    C2 - Alunos deste nível podem entender virtualmente todos os tipos de textos, incluindo escritos abstratos, estruturalmente complexos ou altamente coloquiais, literários e não literários. Eles também podem entender uma grande variedade de textos longos e complexos, apreciando sutis distinções de estilo e significado implícito e explícito.\\

    Provide only the CEFR level as output directly, without explanation or justification.\\
    
    Text: <<\texttt{TEXT}>> \\
    
    Answer:

    \end{tcolorbox}

\begin{figure*}[!t]
    \centering
    \includegraphics[width=\linewidth]{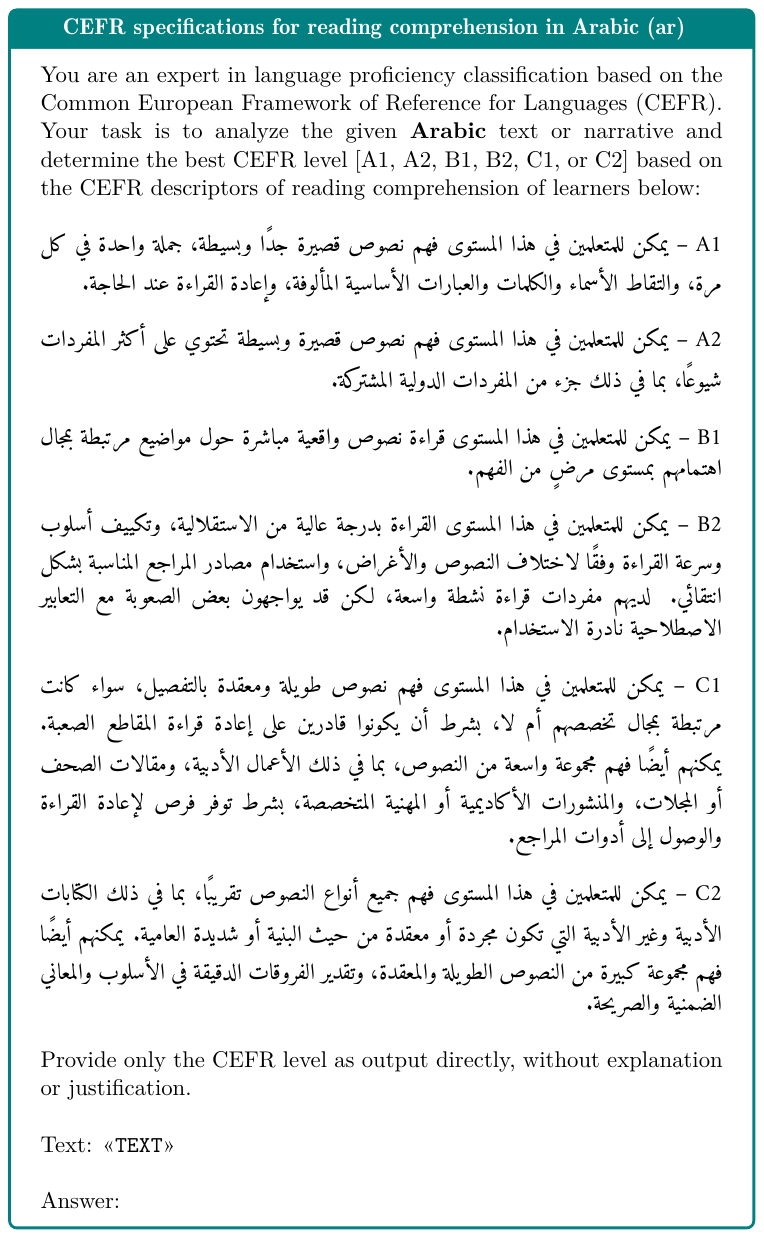}
\end{figure*}
\clearpage

\begin{figure*}[!t]
    \centering
    \includegraphics[width=\linewidth]{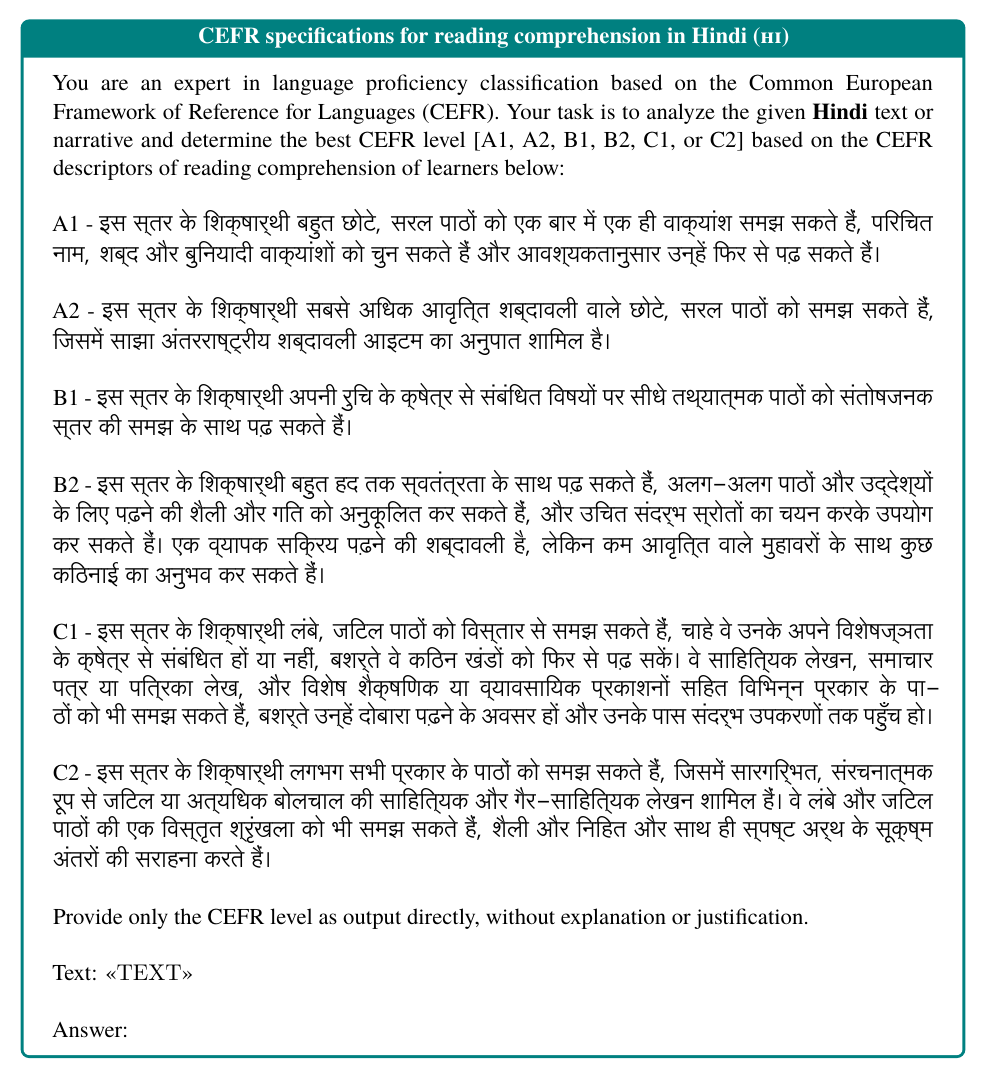}
\end{figure*}

\begin{figure*}[!t]
    \centering
    \includegraphics[width=\linewidth]{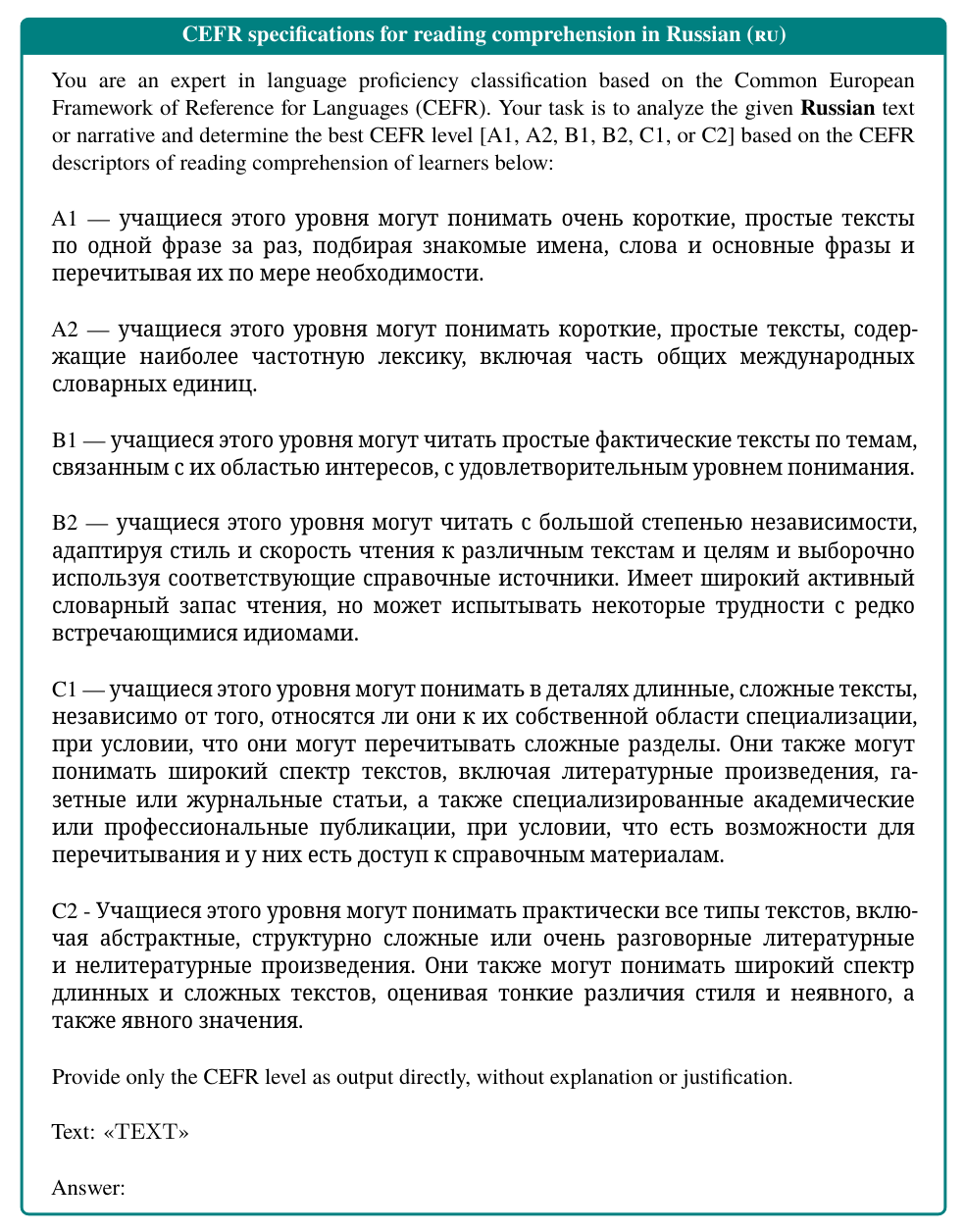}
\end{figure*}
\clearpage

    \begin{tcolorbox}[colframe=teal, colback=white, title=CEFR specifications for reading comprehension in Welsh (\textsc{cy}), coltitle=white, center title, fonttitle=\bfseries]

    You are an expert in language proficiency classification based on the Common European Framework of Reference for Languages (CEFR). Your task is to analyze the given \textbf{Welsh} text or narrative and determine the best CEFR level [A1, A2, B1, B2, C1, or C2] based on the CEFR descriptors of reading comprehension of learners below:\\
    
    A1 - Gall dysgwyr y lefel hon ddeall testunau byr iawn, syml un cymal ar y tro, gan godi enwau, geiriau ac ymadroddion sylfaenol cyfarwydd ac ailddarllen yn ôl yr angen.\\
    
    A2 - Gall dysgwyr y lefel hon ddeall testunau byr, syml sy'n cynnwys yr eirfa fwyaf aml, gan gynnwys cyfran o eitemau geirfa ryngwladol a rennir.\\
    
    B1 - Gall dysgwyr y lefel hon ddarllen testunau ffeithiol syml ar bynciau sy'n ymwneud â'u maes diddordeb gyda lefel foddhaol o ddealltwriaeth.\\
    
    B2 - Gall dysgwyr y lefel hon ddarllen yn annibynnol iawn, gan addasu arddull a chyflymder darllen i wahanol destunau a dibenion, a defnyddio ffynonellau cyfeirio priodol yn ddetholus. Yn meddu ar eirfa ddarllen weithredol eang, ond gall brofi peth anhawster gydag idiomau amledd isel.\\
    
    C1 - Gall dysgwyr y lefel hon ddeall yn fanwl destunau hir a chymhleth, p'un a yw'r rhain yn ymwneud â'u maes arbenigedd eu hunain ai peidio, ar yr amod eu bod yn gallu ailddarllen adrannau anodd. Gallant hefyd ddeall amrywiaeth eang o destunau gan gynnwys ysgrifau llenyddol, erthyglau papur newydd neu gylchgronau, a chyhoeddiadau academaidd neu broffesiynol arbenigol, ar yr amod bod cyfleoedd i'w hail-ddarllen a bod offer cyfeirio ar gael iddynt.\\
    
    C2 - Gall dysgwyr ar y lefel hon ddeall bron bob math o destunau gan gynnwys ysgrifau llenyddol ac anllenyddol haniaethol, strwythurol gymhleth, neu ysgrifau llenyddol ac anllenyddol hynod lafar. Gallant hefyd ddeall ystod eang o destunau hir a chymhleth, gan werthfawrogi gwahaniaethau cynnil o ran arddull ac ystyr ymhlyg yn ogystal ag ystyr amlwg.\\

    Provide only the CEFR level as output directly, without explanation or justification.\\
    
    Text: <<\texttt{TEXT}>> \\
    
    Answer:

    \end{tcolorbox}


    \begin{tcolorbox}[colframe=violet, colback=white, title=CEFR specifications for written production in English (\textsc{en}), coltitle=white, center title, fonttitle=\bfseries]

    You are an expert in language proficiency classification based on the Common European Framework of Reference for Languages (CEFR). Your task is to analyze the given text or narrative and determine the best CEFR level [A1, A2, B1, B2, C1, or C2] based on the CEFR descriptors of reading comprehension of learners below:\\

    A1 - Learners of this level can give information about matters of personal relevance (e.g. likes and dislikes, family, pets) using simple words/signs and basic expressions. Learners can also produce simple isolated phrases and sentences.\\
    
    A2 - Learners of this level can produce a series of simple phrases and sentences linked with simple connectors like “and”, “but” and “because”. Learners have sufficient vocabulary for the expression of basic communicative needs and for coping with simple survival needs.\\
    
    B1 - Learners of this level can produce straightforward connected texts on a range of familiar subjects within their field of interest, by linking a series of shorter discrete elements into a linear sequence. Learners have a good range of vocabulary related to familiar topics and everyday situations.\\
    
    B2 - Learners of this level can produce clear, detailed texts on a variety of subjects related to their field of interest, synthesising and evaluating information and arguments from a number of sources. Learners have a good range of vocabulary for matters connected to their field and most general topics.\\
    
    C1 - Learners of this level can produce clear, well-structured texts of complex subjects, underlining the relevant salient issues, expanding and supporting points of view at some length with subsidiary points, reasons and relevant examples, and rounding off with an appropriate conclusion. Learners can alsoemploy the structure and conventions of a variety of genres, varying the tone, style and register according to addressee, text type and theme.\\
    
    C2 - Learners of this level can produce clear, smoothly flowing, complex texts in an appropriate and effective style and a logical structure which helps the reader identify significant points. Learners have a good command of a very broad lexical repertoire including idiomatic expressions and colloquialisms; shows awareness of connotative levels of meaning.\\
    
    Provide only the CEFR level as output directly, without explanation or justification.\\
    
    Text: <<\texttt{TEXT}>> \\
    
    Answer:

    \end{tcolorbox}

    \begin{tcolorbox}[colframe=violet, colback=white, title=CEFR specifications for written production in Spanish (\textsc{es}), coltitle=white, center title, fonttitle=\bfseries]

    You are an expert in language proficiency classification based on the Common European Framework of Reference for Languages (CEFR). Your task is to analyze the given \textbf{Spanish} text or narrative and determine the best CEFR level [A1, A2, B1, B2, C1, or C2] based on the CEFR descriptors of reading comprehension of learners below:\\
    
    A1 – Los estudiantes de este nivel pueden dar información sobre asuntos de relevancia personal (por ejemplo, gustos y disgustos, familia, mascotas) utilizando palabras/signos simples y expresiones básicas. También pueden producir frases y oraciones simples y aisladas.\\
    
    A2 – Los estudiantes de este nivel pueden producir una serie de frases y oraciones simples conectadas mediante conectores sencillos como “y”, “pero” y “porque”. Tienen un vocabulario suficiente para expresar necesidades comunicativas básicas y afrontar necesidades simples de supervivencia.\\
    
    B1 – Los estudiantes de este nivel pueden producir textos conectados de forma sencilla sobre una variedad de temas conocidos dentro de su campo de interés, enlazando una serie de elementos breves y discretos en una secuencia lineal. Tienen un buen dominio del vocabulario relacionado con temas familiares y situaciones cotidianas.\\
    
    B2 – Los estudiantes de este nivel pueden producir textos claros y detallados sobre diversos temas relacionados con su campo de interés, sintetizando y evaluando información y argumentos de diversas fuentes. Poseen un buen dominio del vocabulario relacionado con su campo y con la mayoría de los temas generales.\\
    
    C1 – Los estudiantes de este nivel pueden producir textos claros y bien estructurados sobre temas complejos, resaltando los aspectos relevantes, desarrollando y respaldando puntos de vista de forma extensa con puntos secundarios, razones y ejemplos pertinentes, y concluyendo adecuadamente. También pueden utilizar la estructura y convenciones de una variedad de géneros, variando el tono, el estilo y el registro según el destinatario, el tipo de texto y el tema.\\
    
    C2 – Los estudiantes de este nivel pueden producir textos claros, fluidos y complejos en un estilo apropiado y efectivo, con una estructura lógica que ayuda al lector a identificar los puntos significativos. Tienen un buen dominio de un repertorio léxico muy amplio, incluyendo expresiones idiomáticas y coloquialismos; muestran conciencia de los niveles connotativos del significado.\\
    
    Provide only the CEFR level as output directly, without explanation or justification.\\
    
    Text: <<\texttt{TEXT}>> \\
    
    Answer:

    \end{tcolorbox}

    \begin{tcolorbox}[colframe=violet, colback=white, title=CEFR specifications for written production in German (\textsc{de}), coltitle=white, center title, fonttitle=\bfseries]

    You are an expert in language proficiency classification based on the Common European Framework of Reference for Languages (CEFR). Your task is to analyze the given \textbf{German} text or narrative and determine the best CEFR level [A1, A2, B1, B2, C1, or C2] based on the CEFR descriptors of reading comprehension of learners below:\\

    A1 – Lernende auf diesem Niveau können Informationen zu persönlich relevanten Themen (z.B. Vorlieben und Abneigungen, Familie, Haustiere) mit einfachen Wörtern/Gebärden und grundlegenden Ausdrücken geben. Sie können auch einfache, isolierte Sätze und Wendungen produzieren.\\
    
    A2 – Lernende auf diesem Niveau können eine Reihe einfacher Sätze und Wendungen bilden, die mit einfachen Konnektoren wie „und“, „aber“ und „weil“ verbunden sind. Sie verfügen über einen ausreichenden Wortschatz, um grundlegende kommunikative Bedürfnisse und einfache Überlebensbedürfnisse auszudrücken.\\
    
    B1 – Lernende auf diesem Niveau können einfache, zusammenhängende Texte zu vertrauten Themen aus ihrem Interessengebiet verfassen, indem sie eine Reihe kürzerer, einzelner Elemente zu einer linearen Folge verknüpfen. Sie verfügen über einen guten Wortschatz zu bekannten Themen und Alltagssituationen.\\
    
    B2 – Lernende auf diesem Niveau können klare, detaillierte Texte zu verschiedenen Themen ihres Interessengebiets verfassen, Informationen und Argumente aus mehreren Quellen zusammenfassen und bewerten. Sie verfügen über einen guten Wortschatz für Themen ihres Fachgebiets sowie für die meisten allgemeinen Themen.\\
    
    C1 – Lernende auf diesem Niveau können klare, gut strukturierte Texte zu komplexen Themen verfassen, die wesentlichen Punkte herausarbeiten, Standpunkte ausführlich mit Nebenaspekten, Begründungen und passenden Beispielen untermauern und mit einem geeigneten Schluss abrunden. Sie können außerdem Aufbau und Konventionen verschiedener Textsorten anwenden und Ton, Stil und Register je nach Adressat, Texttyp und Thema variieren.\\
    
    C2 – Lernende auf diesem Niveau können klare, flüssige und komplexe Texte in einem angemessenen und wirkungsvollen Stil mit einer logischen Struktur verfassen, die dem Leser hilft, wichtige Punkte zu erkennen. Sie beherrschen ein sehr breites Spektrum an Wortschatz, einschließlich idiomatischer Wendungen und umgangssprachlicher Ausdrücke, und zeigen ein Bewusstsein für konnotative Bedeutungsebenen.\\

    Provide only the CEFR level as output directly, without explanation or justification.\\
    
    Text: <<\texttt{TEXT}>> \\
    
    Answer:

    \end{tcolorbox}

    \begin{tcolorbox}[colframe=violet, colback=white, title=CEFR specifications for written production in Dutch (\textsc{nl}), coltitle=white, center title, fonttitle=\bfseries]

    You are an expert in language proficiency classification based on the Common European Framework of Reference for Languages (CEFR). Your task is to analyze the given \textbf{Dutch} text or narrative and determine the best CEFR level [A1, A2, B1, B2, C1, or C2] based on the CEFR descriptors of reading comprehension of learners below:\\

    A1 – Leerlingen op dit niveau kunnen informatie geven over onderwerpen van persoonlijk belang (bijv. voorkeuren en afkeuren, familie, huisdieren) met eenvoudige woorden/tekens en basisuitdrukkingen. Ze kunnen ook eenvoudige, op zichzelf staande zinnen en uitdrukkingen produceren.\\

    A2 – Leerlingen op dit niveau kunnen een reeks eenvoudige zinnen en uitdrukkingen produceren die verbonden zijn met eenvoudige voegwoorden zoals “en”, “maar” en “omdat”. Ze beschikken over voldoende woordenschat om basisbehoeften in communicatie en eenvoudige overlevingssituaties aan te kunnen.\\
    
    B1 – Leerlingen op dit niveau kunnen eenvoudige, samenhangende teksten produceren over een reeks vertrouwde onderwerpen binnen hun interessegebied, door een reeks korte, afzonderlijke elementen in een lineaire volgorde te verbinden. Ze beschikken over een goede woordenschat met betrekking tot vertrouwde onderwerpen en alledaagse situaties.\\
    
    B2 – Leerlingen op dit niveau kunnen duidelijke, gedetailleerde teksten produceren over uiteenlopende onderwerpen die verband houden met hun interessegebied, waarbij ze informatie en argumenten uit meerdere bronnen synthetiseren en evalueren. Ze hebben een goede woordenschat voor onderwerpen binnen hun vakgebied en de meeste algemene thema’s.\\
    
    C1 – Leerlingen op dit niveau kunnen duidelijke, goed gestructureerde teksten produceren over complexe onderwerpen, waarbij ze relevante kernpunten onderstrepen, standpunten uitgebreid onderbouwen met nevenpunten, redenen en relevante voorbeelden, en afsluiten met een passende conclusie. Ze kunnen ook de structuur en conventies van verschillende tekstgenres hanteren en toon, stijl en register aanpassen aan de ontvanger, het teksttype en het thema.\\
    
    C2 – Leerlingen op dit niveau kunnen duidelijke, vloeiende en complexe teksten produceren in een gepaste en effectieve stijl, met een logische structuur die de lezer helpt belangrijke punten te identificeren. Ze beheersen een zeer uitgebreide woordenschat, inclusief idiomatische uitdrukkingen en omgangstaal, en tonen bewustzijn van connotatieve betekenislagen.\\
    
    Provide only the CEFR level as output directly, without explanation or justification.\\
    
    Text: <<\texttt{TEXT}>> \\
    
    Answer:

    \end{tcolorbox}

    \begin{tcolorbox}[colframe=violet, colback=white, title=CEFR specifications for written production in Czech (\textsc{cs}), coltitle=white, center title, fonttitle=\bfseries]

    You are an expert in language proficiency classification based on the Common European Framework of Reference for Languages (CEFR). Your task is to analyze the given \textbf{Czech} text or narrative and determine the best CEFR level [A1, A2, B1, B2, C1, or C2] based on the CEFR descriptors of reading comprehension of learners below:\\

    A1 – Učící se na této úrovni dokážou poskytovat informace o osobně relevantních záležitostech (např. co mají a nemají rádi, rodina, domácí mazlíčci) pomocí jednoduchých slov/znaků a základních výrazů. Dokážou také vytvářet jednoduché izolované fráze a věty.\\
    
    A2 – Učící se na této úrovni dokážou vytvářet sérii jednoduchých frází a vět spojených jednoduchými spojkami jako „a“, „ale“ a „protože“. Mají dostatečnou slovní zásobu pro vyjádření základních komunikačních potřeb a zvládání jednoduchých situací nutných pro přežití.\\
    
    B1 – Učící se na této úrovni dokážou vytvářet přímočaré souvislé texty na řadu známých témat v rámci svého zájmového okruhu, a to spojením série kratších oddělených prvků do lineární posloupnosti. Mají dobrý rozsah slovní zásoby týkající se známých témat a každodenních situací.
    
    B2 – Učící se na této úrovni dokážou vytvářet jasné a podrobné texty o různých tématech souvisejících s jejich oblastí zájmu, přičemž syntetizují a hodnotí informace a argumenty z různých zdrojů. Mají dobrý rozsah slovní zásoby pro témata související s jejich oborem a většinou obecných témat.\\
    
    C1 – Učící se na této úrovni dokážou vytvářet jasné a dobře strukturované texty o složitých tématech, zdůrazňují důležité body, rozvíjejí a podporují názory rozsáhlým způsobem pomocí vedlejších myšlenek, důvodů a relevantních příkladů a zakončují je vhodným závěrem. Také dokážou využívat strukturu a konvence různých žánrů a měnit tón, styl a formálnost podle adresáta, typu textu a tématu.\\
    
    C2 – Učící se na této úrovni dokážou vytvářet jasné, plynulé a složité texty vhodným a efektivním stylem a logickou strukturou, která pomáhá čtenáři rozpoznat důležité body. Mají výbornou znalost velmi široké slovní zásoby včetně idiomů a hovorových výrazů; projevují citlivost na konotace a jemné významové odstíny.\\
    
    Provide only the CEFR level as output directly, without explanation or justification.\\
    
    Text: <<\texttt{TEXT}>> \\
    
    Answer:

    \end{tcolorbox}

    \begin{tcolorbox}[colframe=violet, colback=white, title=CEFR specifications for written production in Italian (\textsc{it}), coltitle=white, center title, fonttitle=\bfseries]

    You are an expert in language proficiency classification based on the Common European Framework of Reference for Languages (CEFR). Your task is to analyze the given \textbf{Italian} text or narrative and determine the best CEFR level [A1, A2, B1, B2, C1, or C2] based on the CEFR descriptors of reading comprehension of learners below:\\
    
    A1 – Gli apprendenti di questo livello possono fornire informazioni su argomenti di rilevanza personale (ad esempio, gusti e preferenze, famiglia, animali domestici) utilizzando parole/segnali semplici ed espressioni di base. Possono anche produrre frasi ed enunciati semplici e isolati.\\
    
    A2 – Gli apprendenti di questo livello possono produrre una serie di frasi ed enunciati semplici collegati con connettivi basilari come “e”, “ma” e “perché”. Possiedono un vocabolario sufficiente per esprimere bisogni comunicativi di base e per affrontare necessità semplici di sopravvivenza.\\
    
    B1 – Gli apprendenti di questo livello possono produrre testi semplici e coerenti su una gamma di argomenti familiari all’interno del proprio campo di interesse, collegando una serie di elementi più brevi in una sequenza lineare. Possiedono un buon repertorio di vocaboli relativi a temi familiari e situazioni quotidiane.\\
    
    B2 – Gli apprendenti di questo livello possono produrre testi chiari e dettagliati su vari argomenti legati al proprio campo di interesse, sintetizzando e valutando informazioni e argomentazioni provenienti da diverse fonti. Hanno un buon vocabolario per trattare argomenti del proprio ambito e la maggior parte dei temi generali.\\
    
    C1 – Gli apprendenti di questo livello possono produrre testi chiari e ben strutturati su argomenti complessi, evidenziando le questioni salienti, sviluppando e sostenendo opinioni in modo articolato con punti secondari, motivazioni ed esempi rilevanti, e concludendo con una chiusura appropriata. Sono anche in grado di adottare la struttura e le convenzioni di diversi generi, variando tono, stile e registro in base al destinatario, al tipo di testo e al tema.\\
    
    C2 – Gli apprendenti di questo livello possono produrre testi chiari, scorrevoli e complessi in uno stile appropriato ed efficace, con una struttura logica che aiuta il lettore a identificare i punti significativi. Possiedono un’ottima padronanza di un ampio repertorio lessicale, incluse espressioni idiomatiche e colloquiali, e mostrano consapevolezza dei livelli connotativi del significato.\\

    Provide only the CEFR level as output directly, without explanation or justification.\\
    
    Text: <<\texttt{TEXT}>> \\
    
    Answer:

    \end{tcolorbox}

    \begin{tcolorbox}[colframe=violet, colback=white, title=CEFR specifications for written production in French (\textsc{fr}), coltitle=white, center title, fonttitle=\bfseries]

    You are an expert in language proficiency classification based on the Common European Framework of Reference for Languages (CEFR). Your task is to analyze the given \textbf{French} text or narrative and determine the best CEFR level [A1, A2, B1, B2, C1, or C2] based on the CEFR descriptors of reading comprehension of learners below:\\

    A1 – Les apprenants de ce niveau peuvent fournir des informations sur des sujets personnels (par exemple, goûts et dégoûts, famille, animaux de compagnie) en utilisant des mots/signes simples et des expressions de base. Ils peuvent également produire des phrases et expressions simples et isolées.\\
    
    A2 – Les apprenants de ce niveau peuvent produire une série de phrases et d’expressions simples reliées par des connecteurs simples comme « et », « mais » et « parce que ». Ils possèdent un vocabulaire suffisant pour exprimer des besoins communicatifs de base et faire face à des situations simples de survie.\\
    
    B1 – Les apprenants de ce niveau peuvent produire des textes clairs et cohérents sur une variété de sujets familiers dans leur domaine d’intérêt, en reliant une série d’éléments plus courts dans une séquence linéaire. Ils disposent d’un bon éventail de vocabulaire lié aux sujets familiers et aux situations de la vie quotidienne.\\
    
    B2 – Les apprenants de ce niveau peuvent produire des textes clairs et détaillés sur divers sujets liés à leur domaine d’intérêt, en synthétisant et en évaluant des informations et arguments issus de plusieurs sources. Ils ont un bon éventail de vocabulaire pour les sujets liés à leur domaine ainsi que pour la plupart des thèmes généraux.\\
    
    C1 – Les apprenants de ce niveau peuvent produire des textes clairs, bien structurés sur des sujets complexes, en soulignant les questions essentielles, en développant et en appuyant leurs points de vue de manière détaillée avec des arguments secondaires, des raisons et des exemples pertinents, et en concluant de manière appropriée. Ils savent aussi utiliser la structure et les conventions de divers genres, en adaptant le ton, le style et le registre selon le destinataire, le type de texte et le thème.\\
    
    C2 – Les apprenants de ce niveau peuvent produire des textes clairs, fluides et complexes dans un style approprié et efficace, avec une structure logique qui aide le lecteur à identifier les points importants. Ils ont une excellente maîtrise d’un très large éventail lexical incluant des expressions idiomatiques et des tournures familières, et font preuve de sensibilité aux niveaux connotatifs de signification.\\
    
    Provide only the CEFR level as output directly, without explanation or justification.\\
    
    Text: <<\texttt{TEXT}>> \\
    
    Answer:

    \end{tcolorbox}

    \begin{tcolorbox}[colframe=violet, colback=white, title=CEFR specifications for written production in Estonian (\textsc{et}), coltitle=white, center title, fonttitle=\bfseries]

    You are an expert in language proficiency classification based on the Common European Framework of Reference for Languages (CEFR). Your task is to analyze the given \textbf{Estonian} text or narrative and determine the best CEFR level [A1, A2, B1, B2, C1, or C2] based on the CEFR descriptors of reading comprehension of learners below:\\

    A1 – Selle taseme õppijad suudavad anda teavet isiklikult olulistel teemadel (nt meeldimised ja mittemeeldimised, perekond, lemmikloomad), kasutades lihtsaid sõnu/viipeid ja põhilisi väljendeid. Õppijad suudavad moodustada ka lihtsaid üksikuid fraase ja lauseid.\\
    
    A2 – Selle taseme õppijad suudavad toota lihtsate fraaside ja lausete jada, mis on seotud lihtsate sidesõnadega nagu „ja“, „aga“ ja „sest“. Neil on piisav sõnavara põhiliste suhtlusvajaduste ja lihtsate ellujäämisvajaduste rahuldamiseks.\\
    
    B1 – Selle taseme õppijad suudavad koostada arusaadavaid, seotud tekste tuttavatel teemadel oma huvivaldkonnas, sidudes lühemaid üksikuid elemente lineaarseks järjestuseks. Neil on hea sõnavara tuttavate teemade ja igapäevaste olukordade kirjeldamiseks.\\
    
    B2 – Selle taseme õppijad suudavad koostada selgeid ja üksikasjalikke tekste erinevatel nende huvivaldkonnaga seotud teemadel, sünteesides ja hinnates teavet ja argumente mitmest allikast. Neil on hea sõnavara oma valdkonnaga seotud teemadeks ning enamike üldiste teemade jaoks.\\
    
    C1 – Selle taseme õppijad suudavad koostada selgeid ja hästi struktureeritud tekste keerukatel teemadel, tuues esile olulised küsimused, laiendades ja toetades seisukohti üksikasjalikult koos täiendavate punktide, põhjuste ja asjakohaste näidetega ning lõpetades sobiva järeldusega. Samuti suudavad nad kasutada erinevate žanrite struktuuri ja konventsioone ning varieerida tooni, stiili ja registrit vastavalt adressaadile, tekstiliigile ja teemale.\\
    
    C2 – Selle taseme õppijad suudavad koostada selgeid, sujuvaid ja keerukaid tekste sobivas ja tõhusas stiilis ning loogilises struktuuris, mis aitab lugejal tuvastada olulisi punkte. Neil on väga lai sõnavara, mis sisaldab idioome ja kõnekeelseid väljendeid; nad tunnetavad ka tähenduse konnotatiivseid tasandeid.\\
    
    Provide only the CEFR level as output directly, without explanation or justification.\\
    
    Text: <<\texttt{TEXT}>> \\
    
    Answer:

    \end{tcolorbox}

    \begin{tcolorbox}[colframe=violet, colback=white, title=CEFR specifications for written production in Portuguese (\textsc{pt}), coltitle=white, center title, fonttitle=\bfseries]

    You are an expert in language proficiency classification based on the Common European Framework of Reference for Languages (CEFR). Your task is to analyze the given \textbf{Portuguese} text or narrative and determine the best CEFR level [A1, A2, B1, B2, C1, or C2] based on the CEFR descriptors of reading comprehension of learners below:\\
    
    A1 – Os aprendentes deste nível conseguem fornecer informações sobre assuntos de relevância pessoal (por exemplo, gostos e preferências, família, animais de estimação) usando palavras/sinais simples e expressões básicas. Também conseguem produzir frases e expressões simples e isoladas.\\
    
    A2 – Os aprendentes deste nível conseguem produzir uma série de frases e expressões simples ligadas por conectores básicos como “e”, “mas” e “porque”. Têm vocabulário suficiente para expressar necessidades comunicativas básicas e lidar com necessidades simples de sobrevivência.\\
    
    B1 – Os aprendentes deste nível conseguem produzir textos simples e coerentes sobre uma variedade de temas familiares dentro de seu campo de interesse, ligando uma série de elementos mais curtos em sequência linear. Possuem um bom repertório de vocabulário relacionado a temas familiares e situações do cotidiano.\\
    
    B2 – Os aprendentes deste nível conseguem produzir textos claros e detalhados sobre uma variedade de assuntos relacionados ao seu campo de interesse, sintetizando e avaliando informações e argumentos de várias fontes. Têm um bom vocabulário para assuntos relacionados à sua área e à maioria dos temas gerais.\\
    
    C1 – Os aprendentes deste nível conseguem produzir textos claros e bem estruturados sobre temas complexos, destacando os pontos relevantes, desenvolvendo e sustentando pontos de vista com argumentos secundários, razões e exemplos pertinentes, e encerrando com uma conclusão apropriada. Também conseguem empregar a estrutura e as convenções de diferentes gêneros textuais, variando o tom, o estilo e o registro conforme o destinatário, o tipo de texto e o tema.\\
    
    C2 – Os aprendentes deste nível conseguem produzir textos claros, fluidos e complexos em um estilo apropriado e eficaz, com uma estrutura lógica que ajuda o leitor a identificar os pontos significativos. Têm um excelente domínio de um repertório lexical muito amplo, incluindo expressões idiomáticas e coloquialismos, e demonstram consciência dos níveis conotativos de significado.\\
    
    Provide only the CEFR level as output directly, without explanation or justification.\\
    
    Text: <<\texttt{TEXT}>> \\
    
    Answer:

    \end{tcolorbox}

\begin{figure*}[!t]
    \centering
    \includegraphics[width=\linewidth]{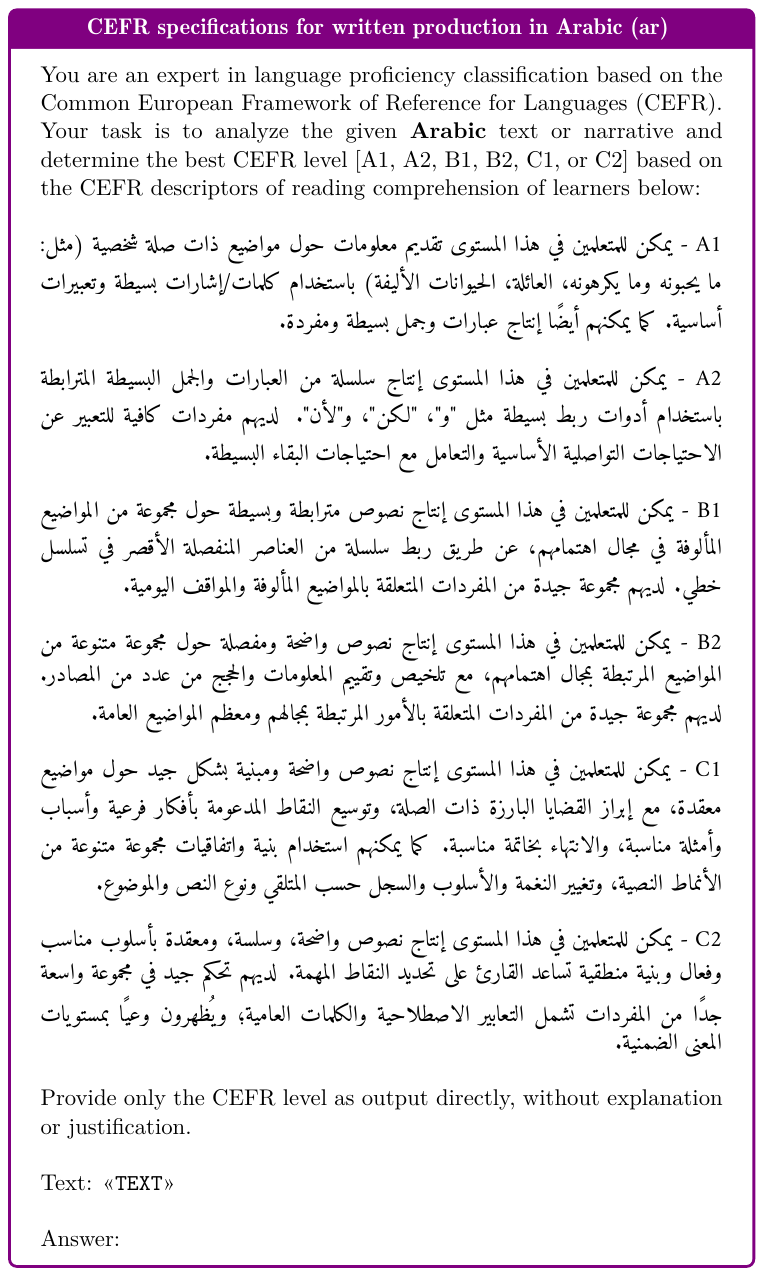}
\end{figure*}
\clearpage

\begin{figure*}[!t]
    \centering
    \includegraphics[width=\linewidth]{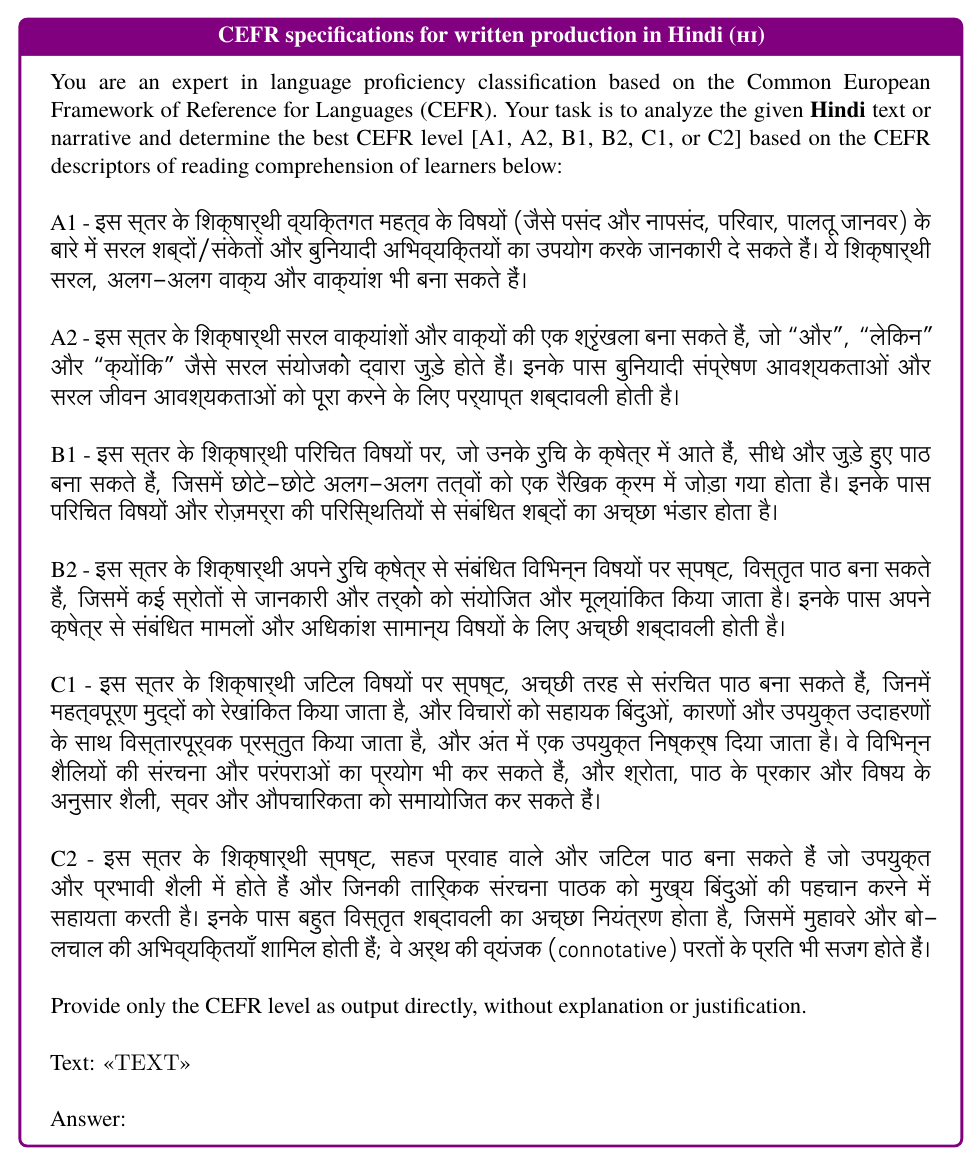}
\end{figure*}

\begin{figure*}[!t]
    \centering
    \includegraphics[width=\linewidth]{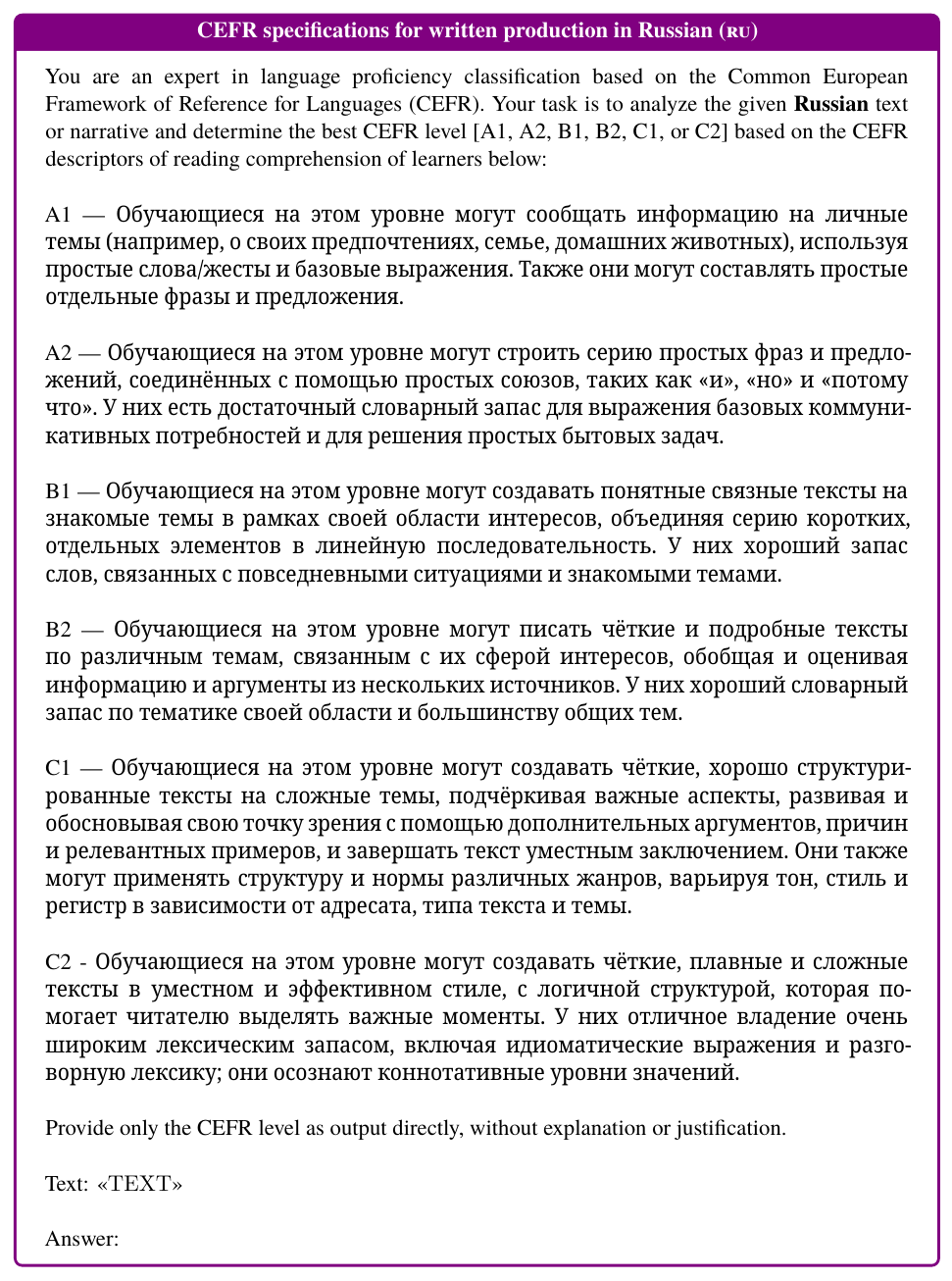}
\end{figure*}
\clearpage

    \begin{tcolorbox}[colframe=violet, colback=white, title=CEFR specifications for written production in Welsh (\textsc{cy}), coltitle=white, center title, fonttitle=\bfseries]

    You are an expert in language proficiency classification based on the Common European Framework of Reference for Languages (CEFR). Your task is to analyze the given \textbf{Welsh} text or narrative and determine the best CEFR level [A1, A2, B1, B2, C1, or C2] based on the CEFR descriptors of reading comprehension of learners below:\\

    A1 – Gall dysgwyr ar y lefel hon roi gwybodaeth am faterion o berthnasedd personol (e.e. pethau maen nhw’n eu hoffi a’u casáu, teulu, anifeiliaid anwes) gan ddefnyddio geiriau/arwyddion syml ac ymadroddion sylfaenol. Gall dysgwyr hefyd gynhyrchu brawddegau ac ymadroddion syml, arwahanol.\\

    A2 – Gall dysgwyr ar y lefel hon gynhyrchu cyfres o ymadroddion a brawddegau syml wedi’u cysylltu gan gysyllteiriau syml fel “a”, “ond” a “oherwydd”. Mae gan ddysgwyr eirfa ddigonol i fynegi anghenion cyfathrebu sylfaenol ac i ymdopi ag anghenion goroesi syml.\\
    
    B1 – Gall dysgwyr ar y lefel hon gynhyrchu testunau cysylltiedig, uniongyrchol ar ystod o bynciau cyfarwydd o fewn eu maes diddordeb, drwy gysylltu cyfres o elfennau byrrach ar wahân i mewn i ddilyniannol linol. Mae ganddynt ystod dda o eirfa sy’n ymwneud â phethau cyfarwydd a sefyllfaoedd bob dydd.\\
    
    B2 – Gall dysgwyr ar y lefel hon gynhyrchu testunau clir, manwl ar amrywiaeth o bynciau sy’n gysylltiedig â’u maes diddordeb, gan gyfuno a gwerthuso gwybodaeth a dadleuon o sawl ffynhonnell. Mae ganddynt ystod dda o eirfa ar gyfer materion sy’n gysylltiedig â’u maes ac ar gyfer y rhan fwyaf o bynciau cyffredinol.\\
    
    C1 – Gall dysgwyr ar y lefel hon gynhyrchu testunau clir, wedi’u strwythuro’n dda ar bynciau cymhleth, gan amlygu’r materion perthnasol, ehangu a chefnogi safbwyntiau’n fanwl gyda phwyntiau ategol, rhesymau ac enghreifftiau perthnasol, a gorffen gyda chasgliad priodol. Gallant hefyd ddefnyddio strwythur a chonfensiynau amrywiaeth o genres, gan amrywio’r naws, arddull a chofrestr yn ôl y derbynnydd, math y testun a’r thema.\\
    
    C2 – Gall dysgwyr ar y lefel hon gynhyrchu testunau clir, esmwyth a chymhleth mewn arddull briodol ac effeithiol ac mewn strwythur resymegol sy’n helpu’r darllenydd i nodi pwyntiau arwyddocaol. Mae ganddynt reolaeth dda dros eirfa eang iawn gan gynnwys ymadroddion idiomatig a llafariad; maent yn dangos ymwybyddiaeth o lefelau ystyron cynhennus.\\

    Provide only the CEFR level as output directly, without explanation or justification.\\
    
    Text: <<\texttt{TEXT}>> \\
    
    Answer:

    \end{tcolorbox}

\end{document}